\documentclass{article}

\newif\ifpreprint
\preprinttrue 

\ifpreprint
  \usepackage[preprint]{style}
\else
  \usepackage{style}
\fi

\usepackage[utf8]{inputenc} 
\usepackage[T1]{fontenc}    
\usepackage{graphicx}       
\usepackage{hyperref}       
\usepackage{url}            
\usepackage{booktabs}       
\usepackage{amsmath}        
\usepackage{amsfonts}       
\usepackage{nicefrac}       
\usepackage{microtype}      
\usepackage{xcolor}         
\usepackage{tikz}           
\usetikzlibrary{arrows.meta,positioning,fit,calc,backgrounds}
\usepackage{xcolor}

\newcommand{\AeroMaybeImage}[3]{%
  \IfFileExists{#1}{\includegraphics[width=#2]{#1}}{%
    \fbox{\parbox[c][#3][c]{#2}{\centering\scriptsize Missing image\\\texttt{\detokenize{#1}}}}%
  }%
}

\title{AeroJEPA: Learning Semantic Latent Representations for Scalable 3D Aerodynamic Field Modeling}

\ifpreprint
  \author{%
    Francisco Giral\thanks{Corresponding authors: fa.giral@alumnos.upm.es; rvinuesa@umich.edu} \\
    Universidad Polit\'ecnica de Madrid
    \And
    Abhijeet Vishwasrao \\
    University of Michigan
    \AND
    Andrea Arroyo Ramo \\
    Universitat Polit\`ecnica de Val\`encia
    \And
    Mahmoud Golestanian \\
    Purdue University
    \And
    Federica Tonti \\
    University of Michigan
    \AND
    Adrian Lozano-Dur\'an \\
    Caltech
    \And
    Steven L. Brunton \\
    University of Washington
    \And
    Sergio Hoyas \\
    Universitat Polit\`ecnica de Val\`encia
    \AND
    Hector Gomez \\ 
    Purdue University
    \And
    Soledad Le Clainche \\
    Universidad Polit\'ecnica de Madrid
    \And
    Ricardo Vinuesa\footnotemark[1] \\
    University of Michigan
  }
\else
  \author{}
\fi

\begin{document}

\maketitle

\begin{abstract} 
{High-fidelity CFD is essential for aerodynamic design, but repeated simulations are computationally expensive, motivating surrogate models for rapid evaluation across geometries and operating conditions. Most existing surrogates are designed for direct field regression, requiring the evaluation of millions of field points even when only an aerodynamic quantity or a localized region is needed, while their internal representations are not intended for direct use in downstream tasks. We introduce \textit{AeroJEPA}, a framework inspired by joint-embedding predictive architectures that represents the problem in two distinct latent spaces: context tokens encode geometry, while predicted tokens encode the aerodynamic state. Both representations remain directly accessible for downstream tasks, such as linear readouts of design variables and aerodynamic quantities without decoding and integrating the full field. When spatial detail is needed, a continuous implicit decoder evaluates the field only at the requested coordinates while reusing the encoded geometry. We evaluate AeroJEPA on HiLiftAeroML, with multi-million-point fields, and SuperWing, which spans a broad family of transonic wings. Compared with state-of-the-art direct-regression surrogates, AeroJEPA trades peak full-field accuracy for compact, reusable representations. In our selective-decoding experiment, however, AeroJEPA substantially outperforms the evaluated direct-regression surrogates while avoiding predictions over the remainder of the aircraft. The learned representations further support controlled interpolation, concept-vector arithmetic, and preliminary constrained latent-space optimization. These results show how predictive representations can support aerodynamic analysis with or without full-field reconstruction. Code is available at \url{https://github.com/NVIDIA/physicsnemo/tree/main}.} 
\end{abstract}

\section{Introduction}

{Aerodynamic design requires repeated evaluation of complex
flow fields across different geometries and operating conditions.
High-fidelity computational fluid dynamics (CFD) remains the primary
tool for obtaining these fields, but its computational cost limits rapid
design-space exploration and optimization. This has long motivated
surrogate modeling for aerodynamic analysis
\citep{Forrester2006Optimization,Forrester2008Engineering,Yondo2017A},
and more recently the development of neural surrogates, neural operators,
and continuous field representations
\citep{Azzizadenesheli2023Neural,Catalani2024Neural,Duvall2025Discretization-independent}.}

{Most existing approaches are designed primarily as direct
geometry-to-field regressors. This is effective when the complete field
is the desired output, but it has two practical limitations. First,
obtaining a localized field or an integrated aerodynamic quantity can
still require predicting millions of values over the complete CFD
discretization. Second, although these models contain intermediate
features, those representations are generally not explicitly organized
or evaluated for reuse in downstream engineering tasks. A model that
instead exposes compact representations of geometry and aerodynamics
could support coefficient prediction, design-variable recovery,
interpolation, and optimization~\citep{vishwasrao2026agentic, vinuesa2026explainable}, without requiring full-field decoding
at every stage. When spatial detail is required, the same representation
could be decoded only at the coordinates relevant to the analysis.}

\begin{figure*}[t]
    \centering
    \includegraphics[width=0.95\textwidth]{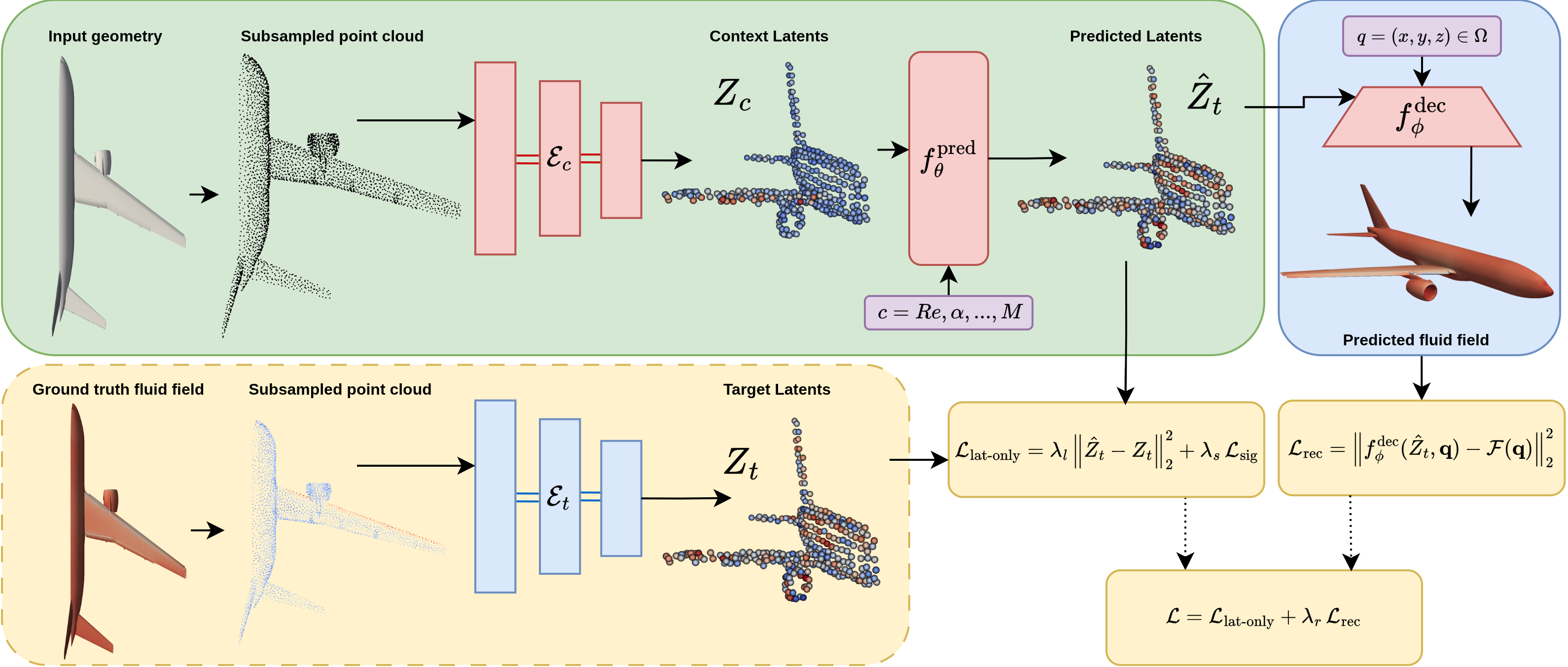}
    \caption{{Overview of AeroJEPA. The context encoder maps
    the geometry point cloud to context tokens $Z_c$, which represent the
    geometry independently of the operating condition. The predictor
    conditions $Z_c$ on the operating variables $c$ (e.g., $\alpha$, $Re$,
    and Mach number) to produce predicted tokens $\hat{Z}_t$ representing
    the aerodynamic state. During end-to-end training, the target encoder
    maps the ground-truth flow to target tokens $Z_t$, providing a
    flow-derived target through $\mathcal{L}_{\mathrm{lat}}$; SIGReg
    $\mathcal{L}_{\mathrm{sig}}$ regularizes the latent distribution, and
    the implicit decoder anchors the predicted tokens to the physical field
    through $\mathcal{L}_{\mathrm{rec}}$. At inference, the target encoder
    is discarded. Context and predicted tokens can be used directly by
    lightweight downstream heads, while the decoder is evaluated only when
    and where field values are requested.}}
    \label{fig:aerojepa_framework}
\end{figure*}

{This objective lies at the intersection of aerodynamic
surrogate modeling and predictive representation learning. Recent
aerodynamic methods have demonstrated compact latent models,
transformer-based surrogates, neural operators, and implicit decoders for
high-dimensional CFD data
\citep{solera2024beta,Solera-Rico2024Towards,eivazi2022towards,
Choy2025Factorized,Wu2024Transolver,Adams2025GeoTransolver,
zou2026adafield}. These methods establish the value of compressed or
continuous field representations, but their primary objective is
typically field reconstruction or direct regression. Joint-Embedding
Predictive Architectures (JEPAs), in contrast, learn by predicting a
target representation from a context representation
\citep{Assran2023Self-Supervised,Sobal2022Joint}. Recent methods such as
LeJEPA and LeWorldModel use explicit latent regularization to stabilize
this process, while related work has extended predictive representation
learning to 3D, multimodal, and scientific settings
\citep{Balestriero2025LeJEPA:,maes2026leworldmodel,Hu2024ThreeDJEPA,
perera2025crossjepa,Chen2025VLJEPA,qu2026representation,Yee2026PIJEPA}.
However, this paradigm has not been developed around the distinct
geometry and aerodynamic representations required for 3D aerodynamic
analysis.}

{We introduce \textit{AeroJEPA}, a framework inspired by
joint-embedding predictive architectures that represents an aerodynamic
problem in two distinct latent spaces. The context encoder produces
tokens describing geometry, while the predictor combines these tokens
with the operating conditions to produce tokens describing the
corresponding aerodynamic state. This separation makes both
representations explicitly accessible: lightweight heads can recover
design variables from geometric information and aerodynamic quantities
from predicted flow information without reconstructing and integrating
the complete field. When local or full-field values are required, a
continuous implicit neural representation (INR) decoder evaluates the
predicted latent state at arbitrary query coordinates. The geometry is encoded
once and its context can be reused across requested coordinates,
subregions, and operating conditions.}

{We evaluate these capabilities on two complementary
datasets. HiLiftAeroML contains high-fidelity boundary-layer fields with
millions of points over realistic high-lift aircraft configurations,
providing a demanding test of large-field and selective decoding.
SuperWing spans a broad parametric family of transonic wings and enables
a detailed evaluation of representation structure and downstream use.
The results expose a deliberate trade-off: compared with direct-regression
surrogates, AeroJEPA can sacrifice peak full-field accuracy in exchange
for compact geometry and flow representations that are directly reusable. This reflects a broader trade-off
between reconstruction fidelity and representation utility; we
therefore evaluate AeroJEPA along both axes rather than solely as a
field regressor. In a
selective-decoding experiment restricted to an outboard-wing region,
however, AeroJEPA outperforms the evaluated direct-regression surrogates
while avoiding predictions over the remainder of the aircraft. Across
the two datasets, the context and predicted representations support
linear readouts, controlled interpolation, concept-vector arithmetic,
and preliminary constrained latent-space optimization.}

{Our contributions are threefold. First, we formulate an
aerodynamic predictive-representation architecture that explicitly
separates geometry context from the condition-dependent aerodynamic
state. Second, we show that these frozen representations support
downstream engineering readouts and structured latent-space operations
without requiring full-field reconstruction. Third, we combine these
representations with a continuous decoder that supports both complete
field reconstruction and selective evaluation at requested coordinates,
and characterize the resulting trade-off against direct-regression
surrogates on two complementary aerodynamic benchmark datasets.}

\section{Method}\label{sec:method}

{AeroJEPA is a predictive representation-learning
framework inspired by joint-embedding predictive architectures and
tailored to 3D aerodynamic problems. It represents the input geometry
and the corresponding aerodynamic state in two distinct token spaces.
The geometry representation can be used directly for geometry-related
downstream tasks, while the predicted aerodynamic representation
supports physical readouts and optional continuous field decoding.}

\subsection{Problem formulation}
\label{subsec:problem_formulation}

We consider the problem of predicting steady-state 3D fluid fields around complex aerodynamic geometries. Let a given geometric design be defined by its boundary surface $\partial \Omega$, which interacts with a fluid over a spatial domain $\Omega \subset \mathbb{R}^3$. The geometry is discretized and represented as an unstructured point cloud of boundary condition (BC) points, $\mathcal{P} = \{x_i\}_{i=1}^{N_c}$, where each $x_i \in \partial \Omega$.

{The flow is governed by a vector of operating or
freestream conditions, denoted by $c$. Depending on the dataset, this
vector may include quantities such as angle of attack $\alpha$,
Reynolds number $Re$, and Mach number. Let
$\mathcal{D}\subseteq\overline{\Omega}$ denote the surface or volumetric
query domain, and let
$\mathcal{F}:\mathcal{D}\rightarrow\mathbb{R}^{d_f}$ denote the
corresponding aerodynamic field, such as pressure, velocity, or
wall-shear quantities.}

{Given $\mathcal{P}$ and $c$, our objective is to learn
two complementary representations: a compact context representation of
the geometry and a predicted representation of the corresponding
aerodynamic state. The geometry representation should support
geometry-related downstream tasks, while the predicted representation
should support aerodynamic readouts without decoding and integrating the
full field. When spatial information is required, the predicted
representation should also retain sufficient information to evaluate
the physical field at any requested query coordinate
$q=(x,y,z)\in\mathcal{D}$.}

\subsection{The AeroJEPA framework}
\label{subsec:aerojepa_framework}

{The core design of AeroJEPA separates geometry encoding,
aerodynamic-state prediction, and field decoding through a tokenized
interface. The context encoder maps the geometry point cloud to context
tokens $Z_c$. Conditioned on the operating variables, the predictor
transforms these tokens into predicted aerodynamic tokens
$\hat{Z}_t$. During training, the target encoder maps the ground-truth
flow to target tokens $Z_t$. At inference, the target encoder is
discarded, leaving the geometry-conditioned path that produces $Z_c$
and $\hat{Z}_t$.}

{The framework admits several possible training strategies. In the
end-to-end configuration used for the primary results, the context
encoder, target encoder, predictor, and decoder are jointly optimized
using latent-matching, latent-regularization, and reconstruction losses.
In a two-stage alternative, inspired by a staged
representation-learning strategy \citep{li2025rethinking}, a flow
autoencoder is first trained for reconstruction; its target encoder and
decoder are subsequently frozen while the context encoder and predictor
learn to match the frozen flow representation from geometry and
operating conditions. A reconstruction-only alternative removes the
target encoder and latent matching and trains the geometry-conditioned
path using field reconstruction as its supervision. A decoder-free configuration could also be formulated
using only latent matching and regularization, but this possibility is
not evaluated in this work. The frozen-target two-stage strategy is
described as an alternative training procedure but is likewise not
included in the experimental comparisons.}

{We use end-to-end training for the primary models because
it learns the geometry representation, aerodynamic representation, and
field decoder in a single optimization procedure. The frozen-target
alternative requires a separate flow-autoencoder pretraining stage
before the geometry-to-latent predictor can be trained, introducing an
additional sequential training cost. End-to-end training also allows the target representation and geometry-conditioned predictor to co-adapt instead
of requiring the predictor to reproduce a fixed latent space. On
HiLiftAeroML, we assess the effect of latent matching through a matched
comparison with reconstruction-only training in
Appendix~\ref{app:loss_ablations}.}

{At inference, context tokens can be used for
geometry-related tasks, while predicted tokens can be used for
aerodynamic readouts conditioned on the operating state. If field
values are required, the predicted tokens are passed to the implicit
decoder, which is evaluated only at the requested coordinates. A
geometry can therefore be encoded once and reused across operating
conditions, downstream readouts, and selective field queries.}

\subsection{Encoders: transforming physical space to latent space}
\label{subsec:encoders}

{AeroJEPA converts irregular geometry and flow point
clouds into fixed-size token sets. This creates a compact interface whose
size does not depend on the number of points in the original CFD field.}

\paragraph{Context Encoder}

{The context encoder, $\mathcal{E}_c$, receives an
unstructured point-cloud representation of the geometry without mesh
connectivity. Each case is first uniformly subsampled to a manageable
input size, typically $8\times10^3$--$131\times10^3$ points depending
on the dataset. Farthest-point sampling (FPS) is then applied within
this input cloud to select spatially distributed tokenizer centers. For surface-only
settings, point coordinates are used as input features. For volumetric
settings, the signed-distance-function value is additionally supplied.
Other geometry-derived attributes such as surface normals could also be incorporated, but were not used during the development of this work.}

Given the subsampled geometry cloud $\mathcal{P}$, the context encoder
produces

\begin{equation}
Z_c=\mathcal{E}_c(\mathcal{P})
\in\mathbb{R}^{M\times d},
\end{equation}

where $M$ is the number of spatial tokens and $d$ is the token width.
In the main experiments, HiLiftAeroML uses $3072$ tokens of width $64$,
while SuperWing uses $512$ tokens of width $128$.

\paragraph{Target Encoder}

{The target encoder, $\mathcal{E}_t$, is used only during
training and operates on an independently sampled point cloud containing
coordinates and ground-truth flow values. Depending on the dataset, this
cloud may contain surface quantities such as $C_p$ and wall-shear
components, or volumetric quantities such as velocity and pressure.
Target points are sampled independently from the geometry input, so the
framework does not require pointwise correspondence between the context
and target point clouds.}

The target encoder produces the flow-derived tokens

\begin{equation}
Z_t=\mathcal{E}_t(\mathcal{F})
\in\mathbb{R}^{M\times d},
\end{equation}

which provide the latent-space target during end-to-end training.

{In both encoders, tokenizer centers are selected by FPS
within the subsampled input cloud. Local neighborhoods around these
centers are aggregated with a lightweight message-passing network
\citep{gilmer2017neural}, and the resulting token set is refined with a
point-transformer backbone. This maps irregular point clouds of varying
size to fixed-size geometry and aerodynamic token sets.}

\subsection{Latent predictor network}\label{subsec:predictor}

The predictive core of AeroJEPA is the predictor network,
$f_{\theta}^{\mathrm{pred}}$, parameterized by $\theta$. Given the
geometry context tokens $Z_c$ and operating conditions $c$, it predicts
the aerodynamic tokens

\begin{equation}
\hat{Z}_t=f_{\theta}^{\mathrm{pred}}(Z_c,c).
\end{equation}

{The operating conditions are injected within the
predictor, allowing one geometry encoding to support multiple operating
states. The prediction stage produces a fixed-size aerodynamic
representation whose cost is independent of the number of field
coordinates subsequently queried. The predicted tokens can be passed
either to lightweight downstream heads or to the field decoder.}

\subsection{Implicit neural representation (INR) decoder}
\label{subsec:inr_decoder}

To map the fluid latent states back into the physical domain, we employ an INR Decoder, $f_{\phi}^{\text{dec}}$, parameterized by $\phi$. Unlike voxel grids or fixed meshes, INRs provide a continuous, resolution-independent mapping by querying exact spatial coordinates.

The decoder takes as input a spatial query point $q = (x, y, z) \in \Omega$ alongside the predicted latents $\hat{Z}_t$. During training, the reconstruction query points are also subsampled from the original field and are chosen independently from both the context and target-encoder point sets. This preserves the decoupled setting and effectively trains the model in a masked fashion, encouraging the latent tokens to encode the underlying continuous field rather than a fixed subset of sampled locations. It outputs the local fluid state variables, such as velocity vector $\mathbf{u}$ and pressure $p$:
\begin{equation}
[\mathbf{u}(q), p(q)] = f_{\phi}^{\text{dec}}(\hat{Z}_t, q)
\end{equation}
By conditioning a multi-layer perceptron (MLP) on the latent vector $\hat{Z}_t$, the INR Decoder acts as a continuous basis function for the aerodynamic field. This architecture is particularly advantageous for complex aircraft geometries, as it completely bypasses the need for volumetric meshing and allows for arbitrary physical resolution purely bounded by the query points provided.

 \subsection{Training objectives}\label{subsec:training_objectives}

{The primary AeroJEPA configuration is trained jointly to
align predicted and flow-derived target tokens, prevent representation
collapse, and retain the ability to reconstruct the physical field.
Following recent JEPA formulations, we use SIGReg
\citep{Balestriero2025LeJEPA:,maes2026leworldmodel} as an explicit
regularizer on the latent distribution.}

For coupled end-to-end training, we optimize

\begin{equation}
\mathcal{L}_{\mathrm{total}}
=
\lambda_{\ell}\mathcal{L}_{\mathrm{lat}}
+
\lambda_r\mathcal{L}_{\mathrm{rec}}
+
\lambda_s\mathcal{L}_{\mathrm{sig}}.
\end{equation}

The latent-matching term is

\begin{equation}
\mathcal{L}_{\mathrm{lat}}
=
\left\|\hat{Z}_t-Z_t\right\|_2^2,
\end{equation}

which aligns the predictor output with the target-encoder output token
by token. The reconstruction term is

\begin{equation}
\mathcal{L}_{\mathrm{rec}}
=
\mathbb{E}_{q\in\mathcal{D}}
\left[
\left\|
f_{\phi}^{\mathrm{dec}}(\hat{Z}_t,q)-\mathcal{F}(q)
\right\|_2^2
\right],
\end{equation}

with the supervised channels determined by the surface or volumetric
prediction task.

{The regularization term
$\mathcal{L}_{\mathrm{sig}}$ applies SIGReg to the latent
representations through random low-dimensional projections toward an
isotropic Gaussian reference distribution. Its role is to discourage
dimensional collapse without requiring an exponential-moving-average
teacher or stop-gradient training heuristic.}

{The primary checkpoints use the coupled end-to-end
objective with $\lambda_{\ell}=1.0$, $\lambda_r=1.0$, and
$\lambda_s=0.001$. These values define the models used in the main
experiments and are not claimed to be universally optimal. On
SuperWing, Appendix~\ref{app:loss_ablations} examines the effect of the
SIGReg weight and a contrastive alternative. On HiLiftAeroML, it
compares end-to-end AeroJEPA with reconstruction-only training of the
same architecture.}

\section{Computational experiments}

{
We evaluate AeroJEPA across two complementary aerodynamic benchmarks, considering full-field accuracy and computational cost, the downstream utility of its geometry and aerodynamic representations, selective decoding, and preliminary latent-space optimization. The selective-decoding protocol, training-objective ablations, and detailed latent-optimization experiment are reported in Appendix~\ref{app:selective_decoding}, Appendix~\ref{app:loss_ablations}, and Appendix~\ref{app:latent_optimization}, respectively.
}

\subsection{Datasets}
We evaluate AeroJEPA on two complementary aerodynamic datasets. The first is the HiLiftAeroML dataset introduced by Ashton et al.~\citep{ashton2026high}, which provides high-fidelity CFD data generated with solution-adapted WMLES for high-lift configurations. This dataset is particularly relevant for assessing whether the proposed latent-space prediction strategy can capture complex separated-flow phenomena on realistic aircraft geometries. The second is SuperWing \citep{yang2025superwing}, a large-scale transonic swept-wing dataset designed for data-driven aerodynamic design. SuperWing contains diverse parameterized wing geometries simulated over a broad range of operating conditions, making it suitable for evaluating generalization across geometry and flow-condition variations. Additional information on the geometry families, flow regimes, and the role of each dataset in our evaluation protocol is provided in Appendix~\ref{app:dataset_details}.

{
We compare AeroJEPA with three established aerodynamic surrogate architectures. FIGConvUNet provides a factorized convolutional approach for learning on high-resolution 3D fields \citep{Choy2025Factorized}, while Transolver is a widely adopted physics-attention model for neural PDE solving \citep{Wu2024Transolver}. GeoTransolver extends this approach with geometry-aware attention and reports state-of-the-art performance on established industrial aerodynamic benchmarks \citep{Adams2025GeoTransolver}. Together, these methods provide strong and complementary direct-regression baselines for evaluating field accuracy, computational cost, and the trade-off between reconstruction fidelity and representation utility.
}

\subsection{HiLiftAeroML: representations and high-resolution field prediction}

\paragraph{Full-field surrogate comparison.}

{
HiLiftAeroML is the most computationally demanding benchmark in our study because it combines realistic high-lift aircraft geometries with approximately $15$ million surface points and $50$ million volumetric points per case. We evaluate the velocity and pressure fields on the boundary-layer surface, where the flow exhibits strong spatial variation. All baseline models are evaluated using their official NVIDIA PhysicsNeMo implementations. For Transolver and GeoTransolver, the surface is divided into randomized chunks of $131{,}072$ points, with every evaluation point processed exactly once to reconstruct the complete evaluated field. The PhysicsNeMo implementation of FIGConvUNet does not support this chunked reconstruction procedure. Consequently, FIGConvUNet predicts only one randomly sampled chunk of $131{,}072$ points from each case, and its reported accuracy and computational cost apply only to that subset rather than to the complete field. AeroJEPA encodes a subsample of $131{,}072$ geometry points, matching the sampling resolution used during training, and predicts a compact aerodynamic latent state. Its implicit decoder is then evaluated over the complete boundary-layer field. The decoded value at a coordinate does not depend on the other query points evaluated in the same batch, allowing the decoder queries to be divided into arbitrary batches without changing the predictions. The baselines receive their prescribed geometric features, including surface normals, whereas AeroJEPA uses point coordinates and operating conditions.

Table~\ref{tab:hilift_full_field} reports relative $L_2$ errors together with floating-point cost, inference time, and model size. All inference measurements were obtained on a single NVIDIA H200 GPU. AeroJEPA matches GeoTransolver in pressure error, obtains the lowest errors for $u$ and $w$, and is second to GeoTransolver for $v$. Reconstructing the complete evaluated field requires $57.00$ TFLOPs and $43$ seconds with AeroJEPA, compared with $189.44$ TFLOPs and $115$ seconds for Transolver and $309.13$ TFLOPs and $187$ seconds for GeoTransolver. AeroJEPA therefore reduces full-field inference time by $62.6\%$ relative to Transolver and by $77.0\%$ relative to GeoTransolver. The $88.34$ TFLOPs and $44$ seconds reported for FIGConvUNet correspond only to one $131{,}072$-point chunk and are not directly comparable with the full-field costs of the other methods. AeroJEPA is also the smallest evaluated model at deployment because the target encoder used during training is discarded at inference.
}

{
\begin{table*}[t]
\centering
\caption{Field-prediction accuracy and computational cost on HiLiftAeroML, reported as mean $\pm$ standard deviation across test cases. Inference times were measured on a single NVIDIA H200 GPU. Transolver, GeoTransolver, and AeroJEPA are evaluated over the complete boundary-layer field. Due to limitations of its official PhysicsNeMo implementation, FIGConvUNet is evaluated on one randomly sampled chunk of $131{,}072$ points per case. Its accuracy, TFLOPs, and inference time are therefore not directly comparable with the full-field results of the other methods. Parameter counts refer to deployed models, except that both training and deployed sizes are reported for AeroJEPA.}
\label{tab:hilift_full_field}
\resizebox{\textwidth}{!}{%
\begin{tabular}{@{}lccccccc@{}}
\toprule
Model &
Rel. $L_2(p)$ $\downarrow$ &
Rel. $L_2(u)$ $\downarrow$ &
Rel. $L_2(v)$ $\downarrow$ &
Rel. $L_2(w)$ $\downarrow$ &
TFLOPs $\downarrow$ &
Time (s) $\downarrow$ &
Parameters (M) $\downarrow$ \\
\midrule
FIGConvUNet
& 0.0066 $\pm$ 0.0009
& 0.2112 $\pm$ 0.0012
& 0.3692 $\pm$ 0.0346
& 0.2538 $\pm$ 0.0337
& 88.34 $\pm$ 2.61
& 44 $\pm$ 1.2
& 6.17 \\
GeoTransolver
& \textbf{0.0048} $\pm$ \textbf{0.0005}
& 0.1503 $\pm$ 0.0017
& \textbf{0.2489} $\pm$ \textbf{0.0287}
& 0.1898 $\pm$ 0.0389
& 309.13 $\pm$ 8.97
& 187 $\pm$ 4.4
& 9.63 \\
Transolver
& 0.0051 $\pm$ 0.0006
& 0.1650 $\pm$ 0.0025
& 0.3180 $\pm$ 0.0289
& 0.1990 $\pm$ 0.0406
& 189.44 $\pm$ 5.50
& 115 $\pm$ 2.5
& 3.99 \\
AeroJEPA (ours)
& \textbf{0.0048} $\pm$ \textbf{0.0026}
& \textbf{0.1445} $\pm$ \textbf{0.0423}
& 0.2749 $\pm$ 0.0388
& \textbf{0.1880} $\pm$ \textbf{0.0455}
& \textbf{57.00} $\pm$ \textbf{1.65}
& \textbf{43} $\pm$ \textbf{0.7}
& \textbf{3.09} deployed / 3.45 training \\
\bottomrule
\end{tabular}%
}
\end{table*}
}

\begin{figure}[t]
    \centering
    \includegraphics[width=0.65\linewidth]{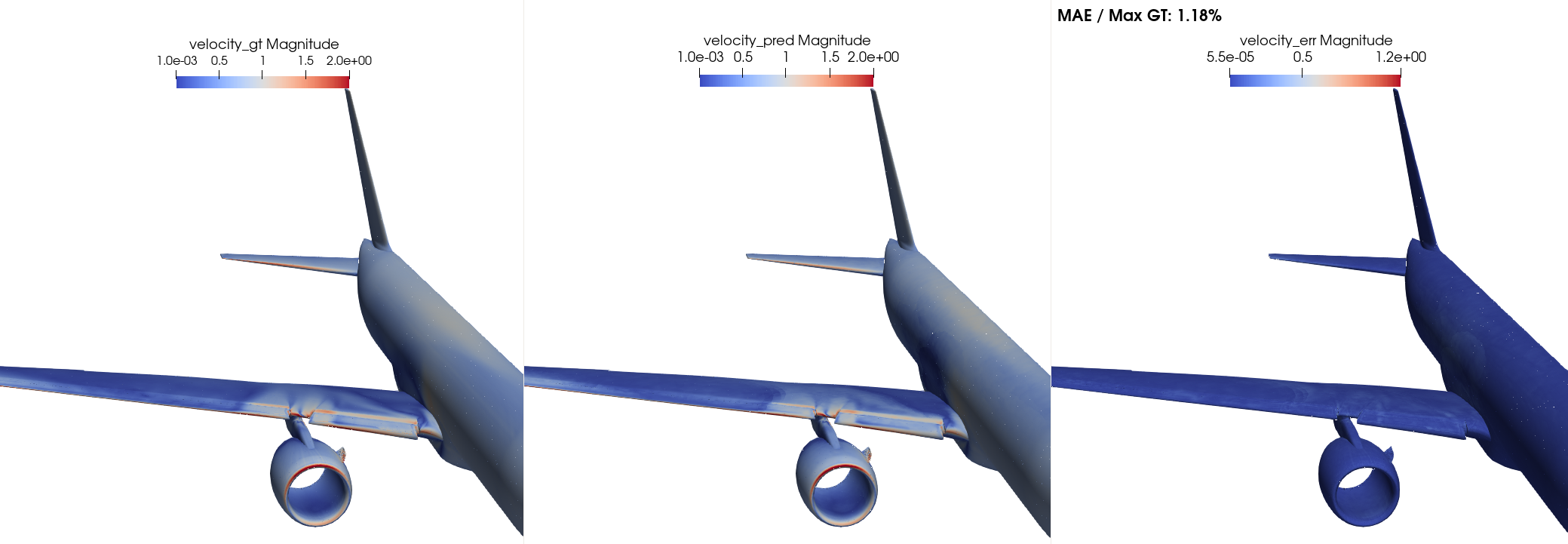}
    \caption{HiLiftAeroML reconstruction for test geometry LHC013 at $18^{\circ}$ angle of attack. AeroJEPA captures the main boundary-layer velocity patterns and high-gradient regions on an unseen configuration while decoding the field continuously over the full aircraft surface. The relative error reported in the figure is $1.18\%$ in MAE / max-GT-mean for the velocity-magnitude field. Additional views are provided in Appendix~\ref{app:highlift_additional}.}
    \label{fig:highlift_case13}
\end{figure}

Beyond aggregate errors, the qualitative reconstructions confirm that the model captures the relevant high-lift flow structures on unseen cases. Figure~\ref{fig:highlift_case13} shows the decoded velocity field for test geometry LHC013 at $18^{\circ}$ angle of attack. The prediction remains consistent with the reference CFD solution in the most sensitive surface regions, including areas affected by strong gradients and flow separation. This qualitative agreement is important because the benchmark evaluates the full resolved boundary-layer field rather than a reduced set of probes or integrated outputs; additional pressure views are provided in Appendix~\ref{app:highlift_additional}.

\begin{figure*}[t]
    \centering
    \begin{minipage}[t]{0.60\textwidth}
        \centering
        \includegraphics[width=\linewidth]{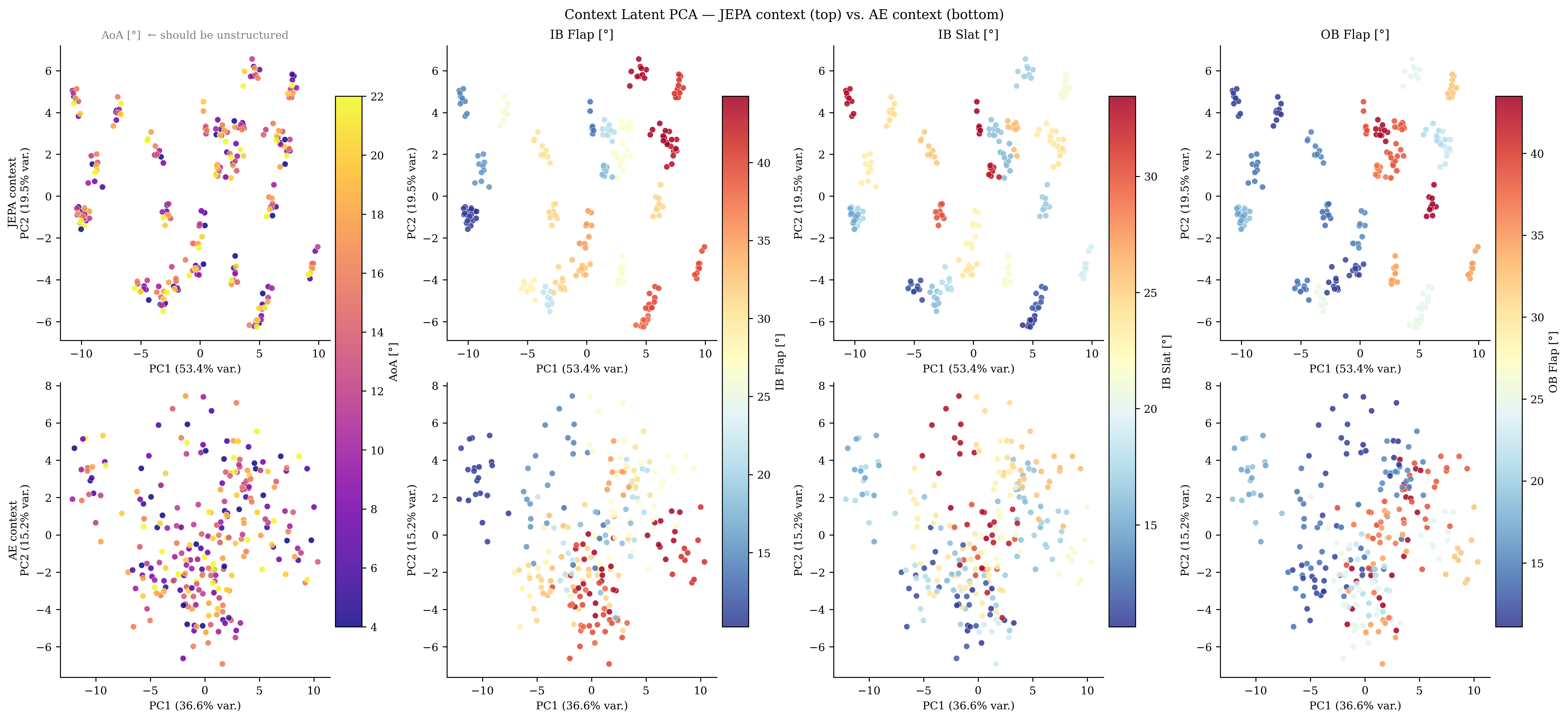}
    \end{minipage}\hfill
    \begin{minipage}[t]{0.60\textwidth}
        \centering
        \includegraphics[width=\linewidth]{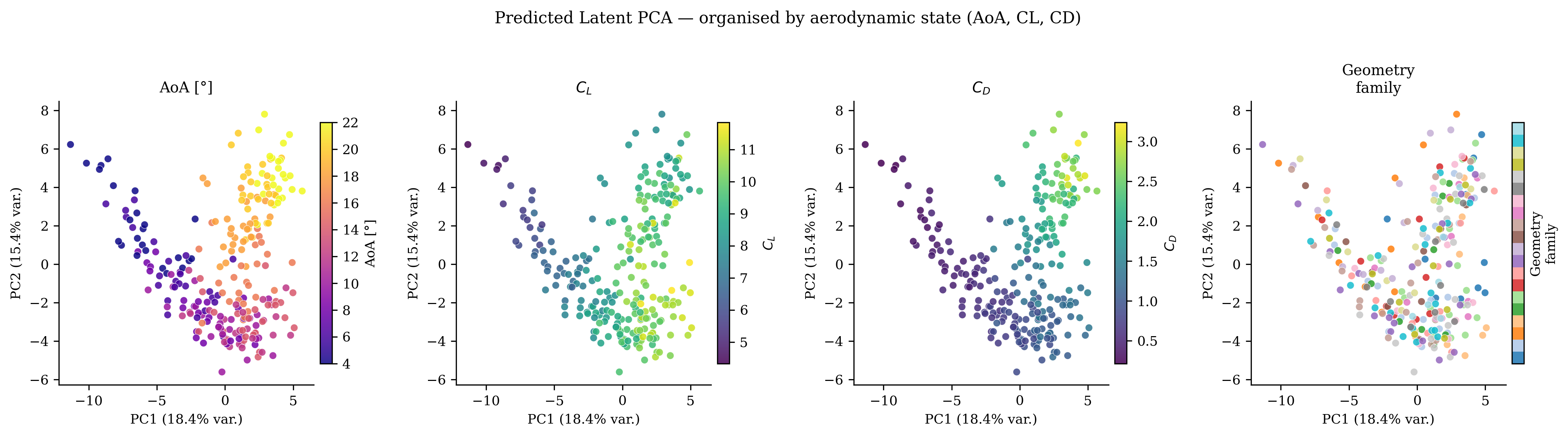}
    \end{minipage}
    {
    \caption{Main HiLiftAeroML latent-space results. \emph{Left:} PCA projection of the context latents comparing AeroJEPA with an autoencoder baseline, showing a more coherent organization by aircraft geometry. \emph{Right:} projection of the predicted latents after conditioning on the operating state, showing smooth variation with aerodynamic quantities even though the model is trained only on the primitive fields $(u,v,w,p)$.}
    }
    \label{fig:highlift_latent_main}
\end{figure*}

\paragraph{Latent analysis.}
We next examine whether the learned latent spaces encode physically meaningful structure. Figure~\ref{fig:highlift_latent_main} compares a PCA projection of the context latents from AeroJEPA against an autoencoder baseline. This baseline preserves a context branch and mixes context and target information through cross-attention before decoding to keep the same information paths as our AeroJEPA, but the training objective is different. AeroJEPA organizes the geometry manifold more coherently, suggesting that the predictive objective induces a more structured representation of the design space. This qualitative separation is supported quantitatively in Appendix~\ref{app:highlift_additional}: ridge probes recover the four dominant high-lift control-surface deflections from the context latent with $R^2$ values between $0.965$ and $0.988$, even though these parameters are never used during training.

The predicted latents after the predictor also encode nontrivial aerodynamic information, despite the fact that the model is trained only on the primitive fields $(u,v,w,p)$ and never on integrated coefficients such as $C_L$ or $C_D$. As shown in Figure~\ref{fig:highlift_latent_main}, the predicted-latent manifold varies smoothly with aerodynamic quantities, and the corresponding linear probes recover $C_L$ and $C_D$ with $R^2=0.939$ and $0.986$, respectively. This indicates that the predictive objective induces a latent representation aligned not only with geometry, but also with downstream aerodynamic performance.

We also find that the context latent is interpretable beyond linear recovery. The concept-vector arithmetic analysis in Appendix~\ref{app:highlift-arithmetic} shows that the main flap and slat directions act as nearly disentangled latent controls: flap-slat cross-talk is weak, while the remaining couplings occur primarily between inboard and outboard elements of the same surface, which is physically consistent with how these controls co-vary in the dataset.

{
Taken together, these experiments show that AeroJEPA combines competitive reconstruction on very large fields with geometry and aerodynamic representations that support probing and controlled latent traversal.
}

{
Appendix~\ref{app:selective_decoding} evaluates selective decoding on an outboard-wing region of 50 held-out cases. AeroJEPA obtains a pressure RMSE of $1.468$ and an $R^2$ of $0.953$, compared with RMSEs of $8.462$ and $8.537$ and $R^2$ values of $-0.161$ and $-0.089$ for Transolver and GeoTransolver, respectively. AeroJEPA also produces identical values when the same coordinates are divided into different query batches, with a repartition NRMSE of zero, compared with $0.0225$ and $0.0134$ for the two direct-regression baselines. These results support localized, query-set-independent decoding.
}

Additional quantitative probe recoveries, interpolation results, and qualitative reconstructions are provided in Appendix~\ref{app:highlift_additional}.

\subsection{SuperWing: broad generalization and latent-space optimization}

{
We use SuperWing to evaluate a complementary regime: generalization across a broad family of transonic swept wings and the usefulness of the learned representations for downstream aerodynamic analysis. Unlike HiLiftAeroML, where the main computational difficulty is the extreme size of each target field, SuperWing contains fixed-resolution surface fields of approximately $32{,}768$ points. The complete field therefore fits in a single forward pass for all evaluated methods, allowing direct comparison with the native one-pass operation of the direct-regression baselines.
}

{
Table~\ref{tab:superwing_full_field} reports full-field relative $L_2$ errors and inference TFLOPs. GeoTransolver and Transolver obtain the lowest reconstruction errors, while AeroJEPA remains more accurate than FIGConvUNet. This is the expected reconstruction-representation trade-off: the direct-regression transformers devote their capacity directly to field accuracy, whereas AeroJEPA additionally organizes the problem into reusable geometry and aerodynamic representations. The SuperWing results therefore do not show an across-the-board reconstruction advantage for AeroJEPA; instead, they establish that the learned representations remain available at a moderate cost in full-field accuracy.
}

{
\begin{table}[t]
\centering
\caption{Full-field relative $L_2$ errors and inference cost on SuperWing, reported as mean $\pm$ standard deviation across test cases. All methods process the complete $32{,}768$-point field in one pass.}
\label{tab:superwing_full_field}
\resizebox{0.82\columnwidth}{!}{%
\begin{tabular}{@{}lcccc@{}}
\toprule
Model &
$C_{f,\tau}$ $\downarrow$ &
$C_{f,z}$ $\downarrow$ &
$C_p$ $\downarrow$ &
TFLOPs $\downarrow$ \\
\midrule
FIGConvUNet
& 0.1429 $\pm$ 0.0525
& 0.2912 $\pm$ 0.1080
& 0.1946 $\pm$ 0.0713
& \textbf{0.27} \\
GeoTransolver
& \textbf{0.0280} $\pm$ \textbf{0.0177}
& \textbf{0.0529} $\pm$ \textbf{0.0335}
& \textbf{0.0309} $\pm$ \textbf{0.0157}
& 0.45 \\
Transolver
& 0.0324 $\pm$ 0.0179
& 0.0606 $\pm$ 0.0341
& 0.0358 $\pm$ 0.0164
& 0.35 \\
AeroJEPA (ours)
& 0.0548 $\pm$ 0.0258
& 0.1084 $\pm$ 0.0513
& 0.0644 $\pm$ 0.0258
& 0.32 \\
\bottomrule
\end{tabular}%
}
\end{table}
}

Additional decoded-field comparisons and force parity plots for SuperWing are reported in Appendix~\ref{app:superwing_additional}.

{
Beyond field reconstruction, we evaluate whether the two latent spaces retain information that can be accessed without decoding and integrating the complete surface field. Linear probes applied to the predicted aerodynamic tokens recover $C_L$ and $C_D$, as well as their within-geometry changes, despite these coefficients not being direct training targets. The geometry tokens similarly retain information about the underlying wing design parameters. Detailed probe results and the loss ablations used to evaluate this representation structure are reported in Appendix~\ref{app:loss_ablations}.
}

{
Finally, we test whether the geometry representation supports preliminary design-space exploration by optimizing the probe-predicted aerodynamic efficiency $C_L/C_D$ at a fixed cruise condition directly in the $128$-dimensional context latent space. The differentiable search uses the frozen AeroJEPA predictor and linear probes for $C_L$ and $C_D$, and is restricted by a latent trust region, bounds on reliably recovered design parameters, aerodynamic constraints, and the observed $L/D$ envelope. The optimization converges toward the high-efficiency region of the dataset without repeatedly decoding the surface field, while the recovered design resembles a high-efficiency wing present in the dataset. Because the synthetic optimum has not been validated with CFD, this result demonstrates constrained exploration of the learned representation rather than improved physical performance. Further details are provided in Appendix~\ref{app:latent_optimization}.
}

{
Taken together, the SuperWing experiments reveal the intended trade-off of the framework. Direct-regression transformers are more accurate full-field predictors on this benchmark, whereas AeroJEPA provides explicit geometry and aerodynamic representations that support lightweight readouts and preliminary latent-space exploration without repeated field decoding. This complements the HiLiftAeroML results, where AeroJEPA remains competitive on multi-million-point fields and permits localized decoding from a reusable full-aircraft geometry representation.
}

\section{Conclusions}\label{sec:conclusions}

{
AeroJEPA is a predictive representation-learning framework for 3D aerodynamics that separates geometry and aerodynamic information into context and predicted tokens, while a continuous decoder provides field values at requested coordinates. Across HiLiftAeroML and SuperWing, these representations support linear readouts of design variables and aerodynamic quantities, controlled interpolation, concept-vector arithmetic, and preliminary latent-space optimization without requiring full-field reconstruction for each downstream task. AeroJEPA remains competitive on the multi-million-point HiLiftAeroML fields with lower computational cost than the evaluated transformer baselines, while direct-regression models achieve greater full-field accuracy on SuperWing. This exposes the intended trade-off between reconstruction fidelity and reusable representations. The selective-decoding experiment further shows that AeroJEPA can evaluate a localized region independently of how its query coordinates are batched.

The present study establishes these capabilities across two complementary steady-flow benchmarks. Future work will extend the framework to unsteady and multi-fidelity regimes, study additional downstream engineering tasks, and integrate the learned representations with inverse-design and shape-parameterization pipelines, including the validation of latent-space design candidates through new CFD simulations.
}


\bibliographystyle{plainnat}
\bibliography{references}

\newpage
\appendix
\section*{Appendix}\label{app}

\section{Other related works}
\label{app:other_related_works}

\paragraph{Broader surrogate-modeling literature.}
Beyond the works emphasized in the main paper, there is a broad literature on surrogate modeling for aerodynamic design, variable-fidelity optimization, and engineering-response approximation \citep{Du2021Rapid,Ghazi2022Training}. There is also extensive work on latent-state and reduced-order surrogate modeling for physical systems more broadly, including latent simulation, data assimilation, and design-space exploration \citep{Hu2022Accelerating,Cheng2022Generalised,Akbari2023Blending,Kontolati2024Learning,Bird2021Dimensionality-reduction-based,Aubeelack2022Surrogate,Zhang2023Heterogeneous}. These papers help situate AeroJEPA within the larger surrogate-modeling landscape, but they are less central to our contribution than methods that explicitly combine aerodynamic field prediction with learned latent structure.

\paragraph{Broader latent-design and JEPA literature.}
On the JEPA side, predictive embedding methods have already been explored for audio, graphs, time series, tabular data, planning, and several scientific domains \citep{Fei2023A-JEPA:,Skenderi2023Graph-level,Verdenius2024LaT-PFN:,Thimonier2024T-JEPA:,Drozdov2024Video,Mo2024Connecting,Mo2024DMT-JEPA:,Garrido2024Learning,Destrade2025Value-guided,Bardhan2025HEP-JEPA:,Katel2024Learning,Chu2026WirelessJEPA:,ElSheikh2026Cell-JEPA,Ulmen2025Learning}. We view these works as evidence that predictive latent learning is broadly useful, while AeroJEPA specializes that paradigm to aerodynamic surrogate modeling with a geometry-to-flow latent bridge.

\section{Additional dataset details}
\label{app:dataset_details}

This section summarizes the two datasets utilized throughout the paper, emphasizing geometry parameterization, observable fluid variables, and the structural point-cloud representation mandated by the AeroJEPA pipeline. Furthermore, we outline the exact dataset partition strategies formulated to rigorously evaluate generalizability.

\subsection{HiLiftAeroML}
HiLiftAeroML \citep{ashton2026high} serves as the high-fidelity aerodynamic benchmark in our study. The dataset catalogs realistic high-lift aircraft configurations simulated via solution-adapted wall-modeled large-eddy simulation (WMLES). This physical complexity makes it particularly relevant for evaluating continuous latent representations under highly separated, turbulent flow. Computationally, it represents the most demanding environment from a spatial scale perspective. In our formulation, the surface-oriented boundary-layer representation resolves approximately $12$--$15$ million points per sample, whereas the internal volumetric representation constructs roughly $50$ million elements per case.

Geometrically, the configurations are denoted by specific design geometry (e.g., "LHCXX") defined internally by 8 continuous design parameters. Four of these scalars dictate the highest observed flow variation: the inboard and outboard flap deflections alongside the inboard and outboard slat deflections. For every given geometry configuration, the dataset provides $10$ discrete operational flow snapshots sweeping across an Angle of Attack (AoA) domain bounding from $4^\circ$ to $22^\circ$. Given the extreme computational burden intrinsic to evaluating millions of points across full WMLES traces, we strictly partition the benchmark topologies into $205$ cases reserved for the training manifold, and $50$ cases held out for testing.

\subsection{SuperWing}
SuperWing \citep{yang2025superwing} acts as the large-scale transonic benchmark designed to evaluate continuous generalization across topological and operational variabilities. The corpus contains exactly $4{,}239$ parameterized wing geometries mapped across $28{,}856$ unique Reynolds-averaged Navier--Stokes (RANS) state solutions spanning a broad aerodynamic domain of Mach numbers and freestream angles of attack. Each simulation is originally distributed on a structured surface discretization of roughly $32{,}000$ points. To ensure architectural parity, we treat these grids purely as unstructured point sets so the AeroJEPA pipeline remains invariant across dataset discretizations. 

Target fluid fields across SuperWing are iteratively mapped as continuous surface vectors, evaluating the local pressure coefficient ($C_p$) and skin-friction coefficient ($C_f$), contrary to volumetric arrays. Geometrically, each surface is prescribed by $54$ latent morphological degrees of freedom regulating localized shape, wing twist, and spanwise dihedral. 

Crucially, to rigorously enforce zero-shot geometric generalizability, the dataset partitioning is structured strictly across the geometrical dimension rather than by random simulation sweeps. The parameterized wings are isolated into an $80\%/10\%/10\%$ split corresponding to training, validation, and testing regimes, respectively. Because the stratification dictates that all unique operating parameters for a specific held-out wing topology remain completely unseen during optimization, the evaluation cleanly measures fundamental physical extrapolation over unseen boundary shapes.

\section{Additional model details}
\label{app:model_details}

This appendix provides a comprehensive, step-by-step description of the AeroJEPA pipeline, ranging from raw computational fluid dynamics (CFD) representation to latent prediction and subsequent continuous field decoding.

\subsection{Point-cloud preprocessing and subsampling}
Standard unstructured CFD meshes exhibit highly variable node densities. To standardize the input representation, connectivity operations are discarded, allowing both the physical boundary and the surrounding flow to be formally treated as continuous point sets. Let the complete dense surface geometry be denoted as $\mathcal{P}$, and the comprehensive ground-truth fluid states over the spatial domain $\Omega$ as $\mathcal{F}$. For surface-only fluid problems, the geometry branch processes purely spatial coordinates \(\mathbf{x} \in \mathbb{R}^3\). For volumetric domains, this input is strictly augmented with the local Signed-Distance Function (SDF). 

Because the complete CFD point sets exceed the memory required for direct processing, we uniformly sample bounded observation sets for each case. The geometry branch receives a context set $\mathcal{P}_c\subset\mathcal{P}$ of cardinality $N_c$, while the target branch receives an independently sampled flow set $\mathcal{F}_t\subset\mathcal{F}$ of cardinality $N_t$. A third independently sampled coordinate set $\mathcal{Q}\subset\Omega$, with cardinality $N_q$, provides the reconstruction queries. Farthest point sampling is applied only after this initial sampling step to select the tokenizer centroids within each bounded set.

The geometry, target, and reconstruction sets are sampled independently, so no pointwise correspondence is imposed between their elements. Incidental overlap is not prohibited, but the decoder queries are not aligned with the context or target tokens. This separation prevents the model from relying on a fixed pointwise mapping and trains the predicted tokens to support field evaluation at independently selected coordinates.

\subsection{Tokenization with sampled centroids}
Following subsampling, the irregular target and context point sets are projected into fixed-length sequences of latent spatial tokens. Specifically, the architecture leverages FPS once more over the subsampled sets $\mathcal{P}_c$ and $\mathcal{F}_t$ to designate foundational centroid coordinates. A localized message-passing neighborhood aggregation layer compiles point-wise variables around each centroid into preliminary latent tokens. This projection efficiently transitions the variable-scale physical point cloud into a structured, bounded topology compatible with global transformer operations. Based on the respective domain complexities, the token context sequence lengths ($M$) and feature dimensions ($d$) are scaled; the high-dimensional spatial topologies resident in HiLiftAeroML necessitate sequences of $M=3072$ tokens wherein $d=64$, whereas the lower-resolution surface fields intrinsic to SuperWing are sufficiently parameterized by $M=512$ tokens of dimensionality $d=128$.

\subsection{Encoder backbones}
Both the context encoder $\mathcal{E}_c$ and the target encoder $\mathcal{E}_t$ utilize computational graphs built upon localized Point Transformer attention mechanisms \citep{wu2024point}. Crucially, the massive dimensionality reduction separating the raw $N_c$ and $N_t$ inputs from the computational bounds is handled entirely by the preceding centroid-clustering and neighborhood aggregation layers. Consequently, the encoder backbones do not inherently alter the topological scale; they operate purely as flat processing stacks over the highly compressed, $M$-sized token distributions. By employing localized self-attention rather than global quadratic computations, the framework strictly bounds memory consumption. Through progressively stacked blocks, the encoder incrementally routes physical information across neighboring tokens. Shallow layers capture highly localized boundary layer interactions and surface curvature semantics, whereas deeper transformer blocks contextualize macro-aerodynamic coupling and wake propagation over multiple token hops. The resulting latent interface inherently preserves the geometric topology, formulating an optimizable blueprint for generative design.

\subsection{Predictor architecture and conditioning}
Sharing structural topology with the encoders, the core latent map is represented by a non-hierarchical predictor network operating seamlessly over the fixed output token resolution. The objective of the predictor is to synthesize target flow tokens purely from the geometry context tokens, and the driving physical constraints, without requiring spatial pooling operations. 

Architecturally, the predictor instantiates learnable latent queries corresponding to the spatial coordinates of a fixed set of centroids. These spatial query coordinates are first embedded via high-frequency Fourier positional encodings. The network processes these queries through a stacked, flat sequence that tightly intertwines localized point-transformer self-attention and token-based cross-attention mechanisms. The self-attention pathway facilitates continuous spatial refinement and consistency among the target queries themselves, whereas the interleaving cross-attention layers allow these queries to explicitly fetch aligned geometric features from the upstream context tokens acting as keys and values.

Furthermore, recognizing that minor perturbations in freestream operating variables (e.g., Angle of Attack $\alpha$, Reynolds Number $Re$, Mach Number $Ma$) fundamentally bifurcate the boundary layer dynamics independent of structural modification, we inject these physics conditions natively into the network. A lightweight multi-layer perceptron processes the scalar variables, projecting them deeply into the hidden transformer layers through adaptive feature modulation \citep{peebles2023scalable}. Consequently, the predictor framework robustly maps an invariant geometric parameterization into a variable continuous flow topology, establishing a smooth, differentiable property leveraged subsequently by gradient-based latent optimization probes.

\subsection{INR decoder implementation details}
To remap the predicted token distribution to an arbitrarily resolved continuous space, the predicted tokens $\hat{Z}_t$ condition the INR decoder. For any arbitrary element $\mathbf{q}_i \in \mathcal{Q}$, the coordinate is first transformed via Fourier feature encodings to enhance high-frequency spatial sensitivity before being processed by the decoding multi-layer perceptron (MLP). In volumetric settings, query elements are augmented by the respective SDF evaluations, explicitly distinguishing the interior solid boundary, the near-wall viscous boundary layer, and the external flow.

\subsection{Training configurations and hyperparameters}
The primary models are trained end to end with latent-matching and reconstruction weights $\lambda_{\ell}=1.0$ and $\lambda_r=1.0$, and SIGReg weight $\lambda_s=0.001$. This configuration provides the strongest overall reconstruction and representation results in the SuperWing ablation reported in Appendix~\ref{app:loss_ablations}. Network weights are optimized using AdamW with gradient clipping. All models, including the baselines, were trained on a single NVIDIA H200 GPU. On HiLiftAeroML, AeroJEPA required approximately $48$ hours for $300$ epochs, while the baseline models required approximately $96$ hours for the same number of epochs. On SuperWing, AeroJEPA and the baseline models required approximately $24$ hours for $200$ epochs. Table~\ref{tab:hyperparameters} summarizes the sampling cardinalities, architectural dimensions, and optimization settings.

\begin{table}[h]
\centering
\caption{Primary architectural parameterization and optimization bounds for the AeroJEPA framework, delineated by the underlying aerodynamic environment.}
\label{tab:hyperparameters}
\begin{tabular}{lcc}
\toprule
\textbf{Configuration Variable} & \textbf{HiLiftAeroML} & \textbf{SuperWing} \\ 
\midrule
\multicolumn{3}{c}{\textit{Sampling Capacities}} \\
Geometry Context Subsample ($N_c$) & $131,072$ & $8,192$ \\
Fluid Target Subsample ($N_t$) & $131,072$ & $8,192$ \\
Reconstruction Query Set ($N_q$) & $131,072$ & $8,192$ \\
\midrule
\multicolumn{3}{c}{\textit{Latent Space Structure}} \\
Context Tokens ($M$) & $3,072$ & $512$ \\
Token Dimensionality ($d$) & $64$ & $128$ \\
Transformer Backbone Depth & $6$ Layers & $6$ Layers \\
\midrule
\multicolumn{3}{c}{\textit{Optimization Setup}} \\
Optimizer Formulation & AdamW & AdamW \\
Learning Rate & $1 \times 10^{-3}$ & $1 \times 10^{-3}$ \\
LR Annealing Schedule & Cosine Warmup & Cosine Warmup \\
Weight Decay & $1 \times 10^{-3}$ & $1 \times 10^{-3}$ \\
Gradient Clip Threshold & $10.0$ & $10.0$ \\
Total Training Epochs & $300$ & $200$ \\
\bottomrule
\end{tabular}
\end{table}

\section{Additional results and experiments}
\label{app:additional_results}

\subsection{Additional HiLiftAeroML results}
\label{app:highlift_additional}

\begin{figure*}[t]
    \centering
    \includegraphics[width=0.8\textwidth]{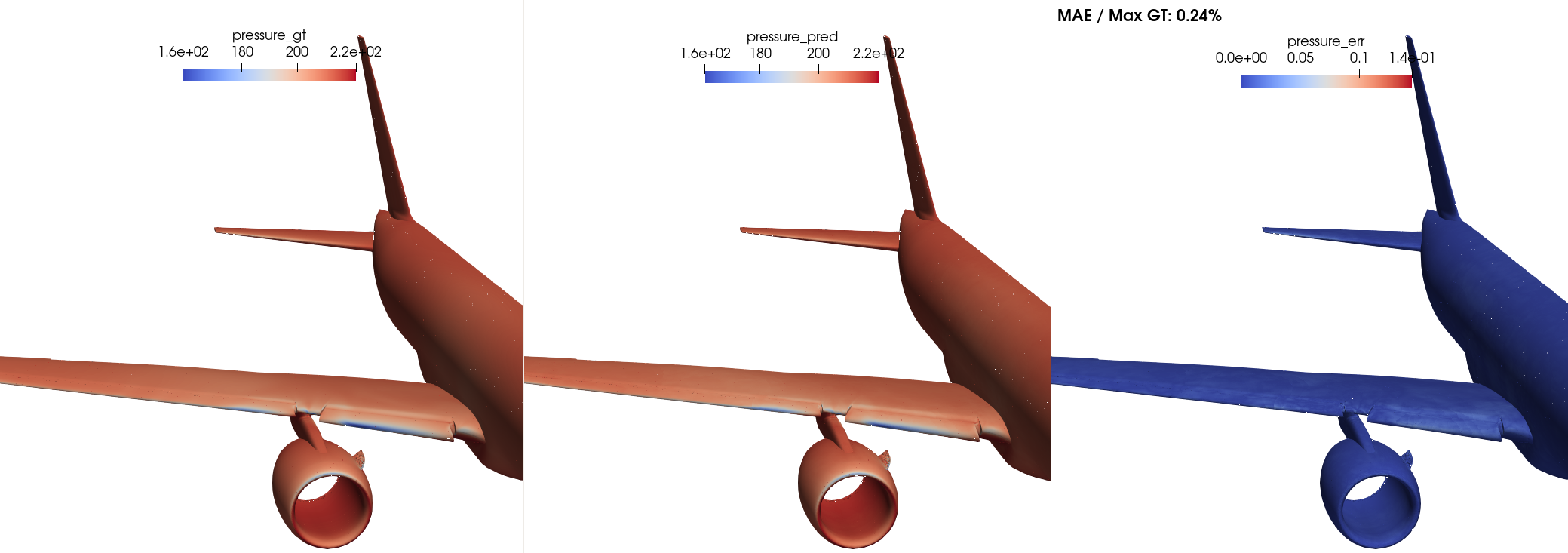}\hfill
    \includegraphics[width=0.8\textwidth]{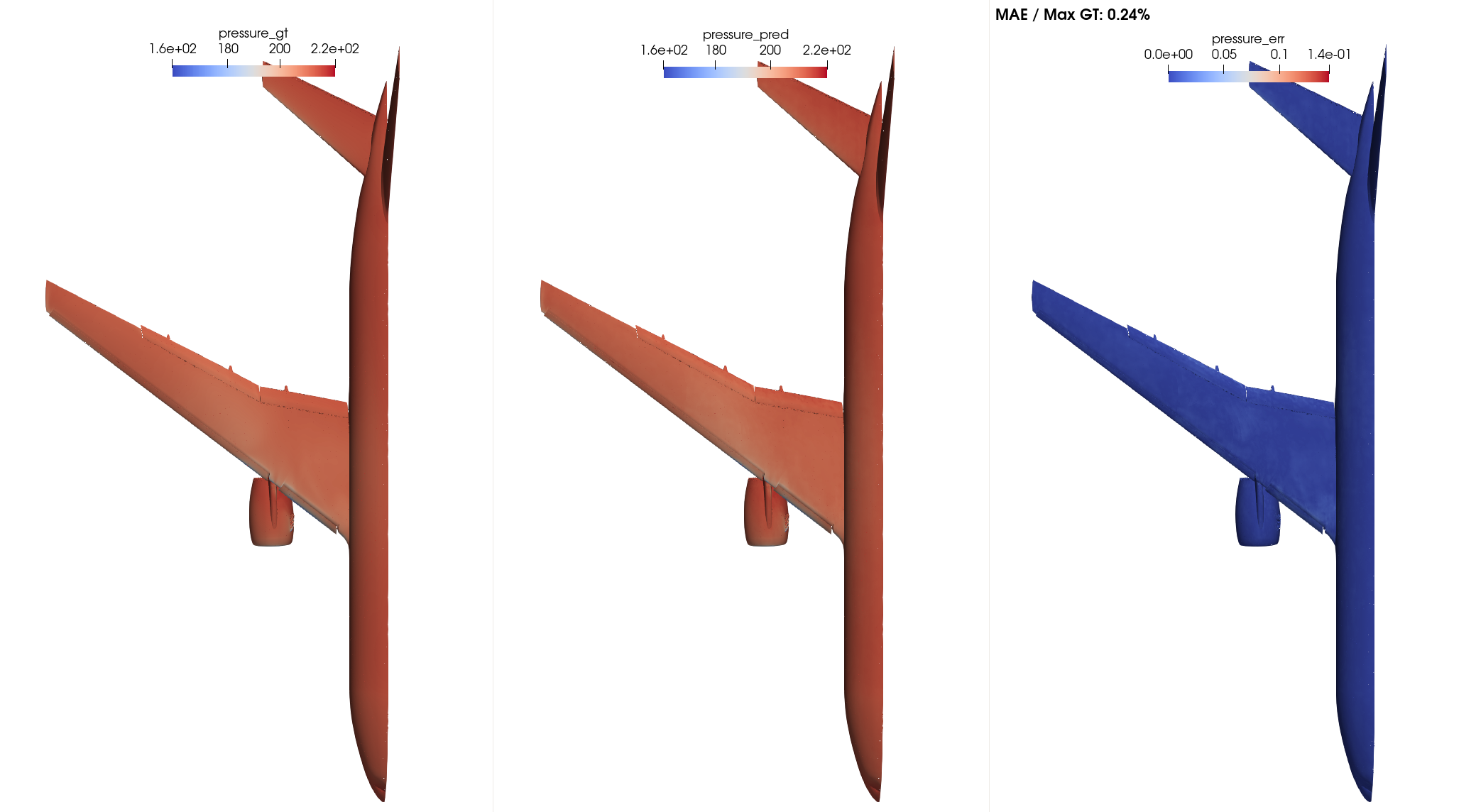}
    \caption{Additional qualitative pressure views for test geometry LHC013 at $18^{\circ}$ angle of attack. These complementary views support the velocity-based qualitative comparison shown in the main text and confirm that AeroJEPA preserves the main surface-pressure structure on the unseen case. The relative error reported in the figure is $0.24\%$ in MAE / max-GT-mean for the pressure field.}
    \label{fig:highlift_case13_appendix}
\end{figure*}

\paragraph{Qualitative reconstruction.}
Figure~\ref{fig:highlift_case13_appendix} gathers additional qualitative views of the decoded fields for test geometry LHC013 at $18^{\circ}$ angle of attack. Together with the velocity view retained in the main text, these complementary pressure and velocity plots confirm the same pattern: AeroJEPA preserves the large-scale structure of the boundary-layer solution and remains consistent with the reference CFD field in challenging regions.

\begin{figure*}[t]
    \centering
    \includegraphics[width=0.72\textwidth]{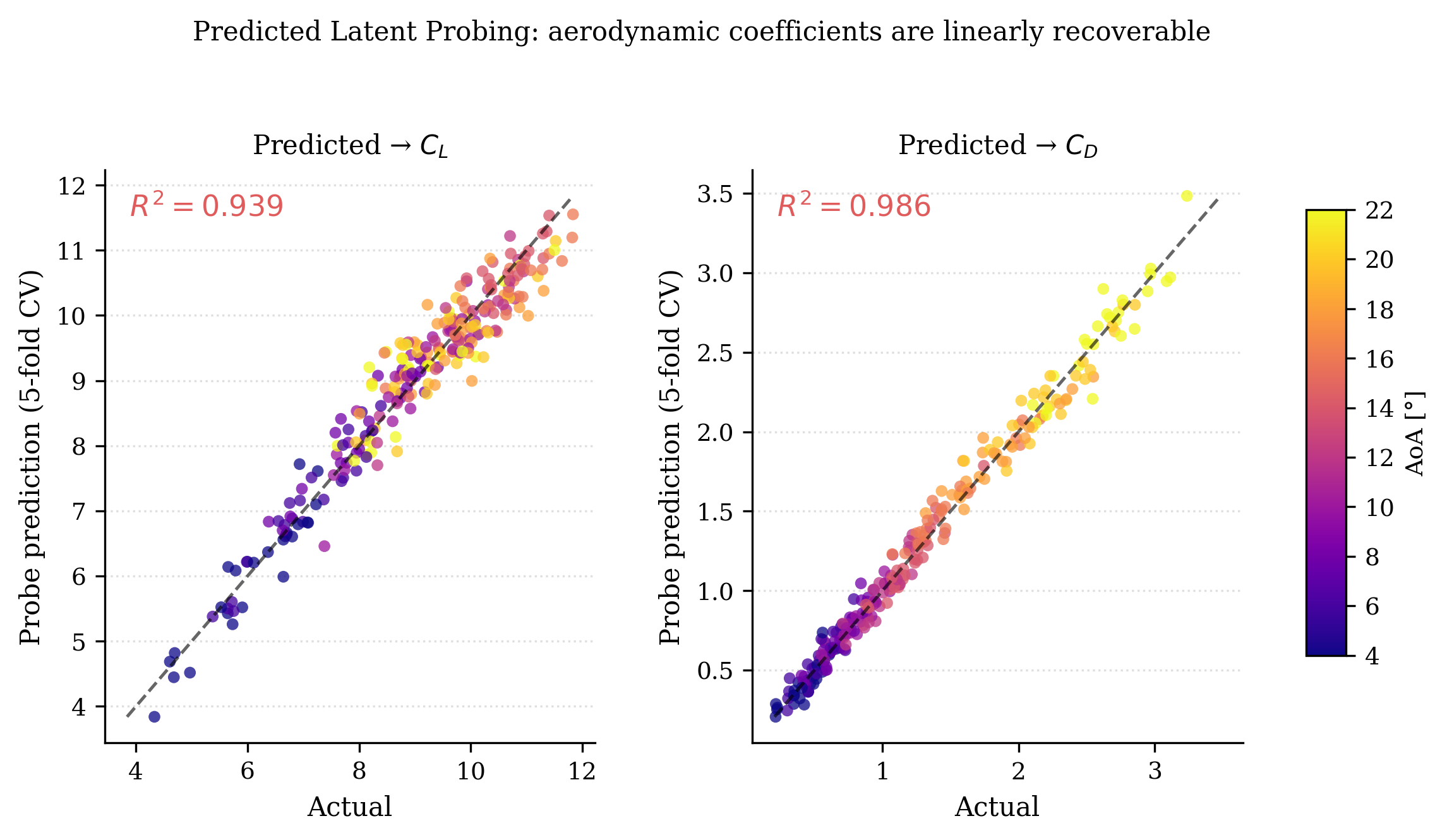}
    \caption{Ridge linear probing from predicted latents to aerodynamic coefficients on HiLiftAeroML. Even though the model is trained only on the primitive fields $(u,v,w,p)$, the predicted latent supports accurate recovery of $C_L$ and $C_D$, consistent with the smooth organization of the predicted manifold shown in the main text.}
    \label{fig:highlift_probe_appendix}
\end{figure*}

\begin{figure*}[t]
    \centering
    \includegraphics[width=0.82\textwidth]{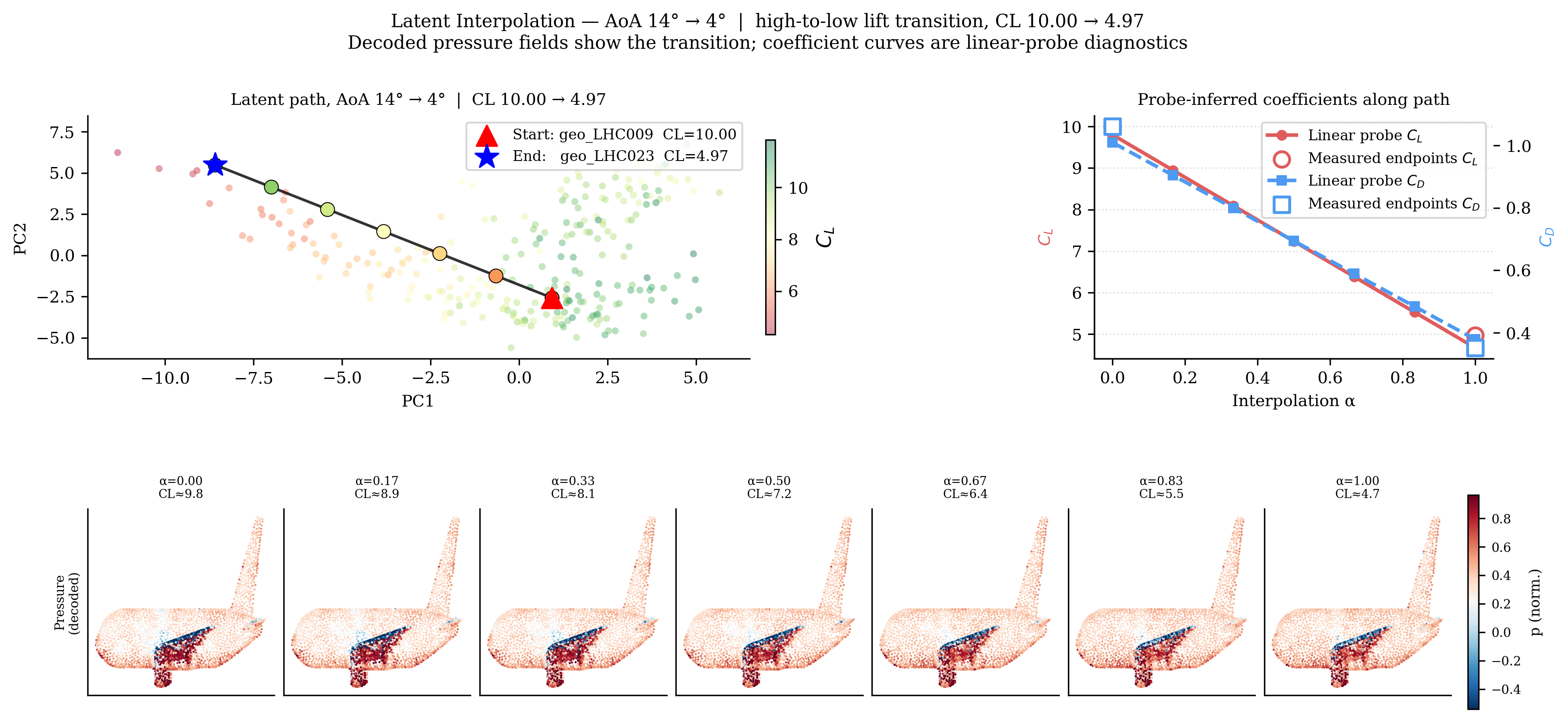}
    \caption{Latent interpolation on HiLiftAeroML between different operating conditions and geometry states. AeroJEPA yields decoded fields that remain physically meaningful across the interpolation path, with integrated coefficients that stay close to the corresponding ground-truth values.}
    \label{fig:highlift_interpolation_appendix}
\end{figure*}

\paragraph{Latent organization and interpolation.}
Figure~\ref{fig:highlift_probe_appendix} reports the ridge-probe recovery of aerodynamic coefficients from the predicted latent, complementing the predicted-manifold visualization shown in the main text. Figure~\ref{fig:highlift_interpolation_appendix} then examines interpolation in the latent space. We introduce a scalar interpolation parameter $\alpha$ and move between latent states corresponding to different angles of attack and geometry configurations. The decoded fields remain physically meaningful along the trajectory, and the resulting aerodynamic coefficients stay close to the ground-truth values, indicating that the learned manifold supports smooth transitions across both operating conditions and design changes.

\begin{figure*}[t]
    \centering
    \includegraphics[width=0.72\textwidth]{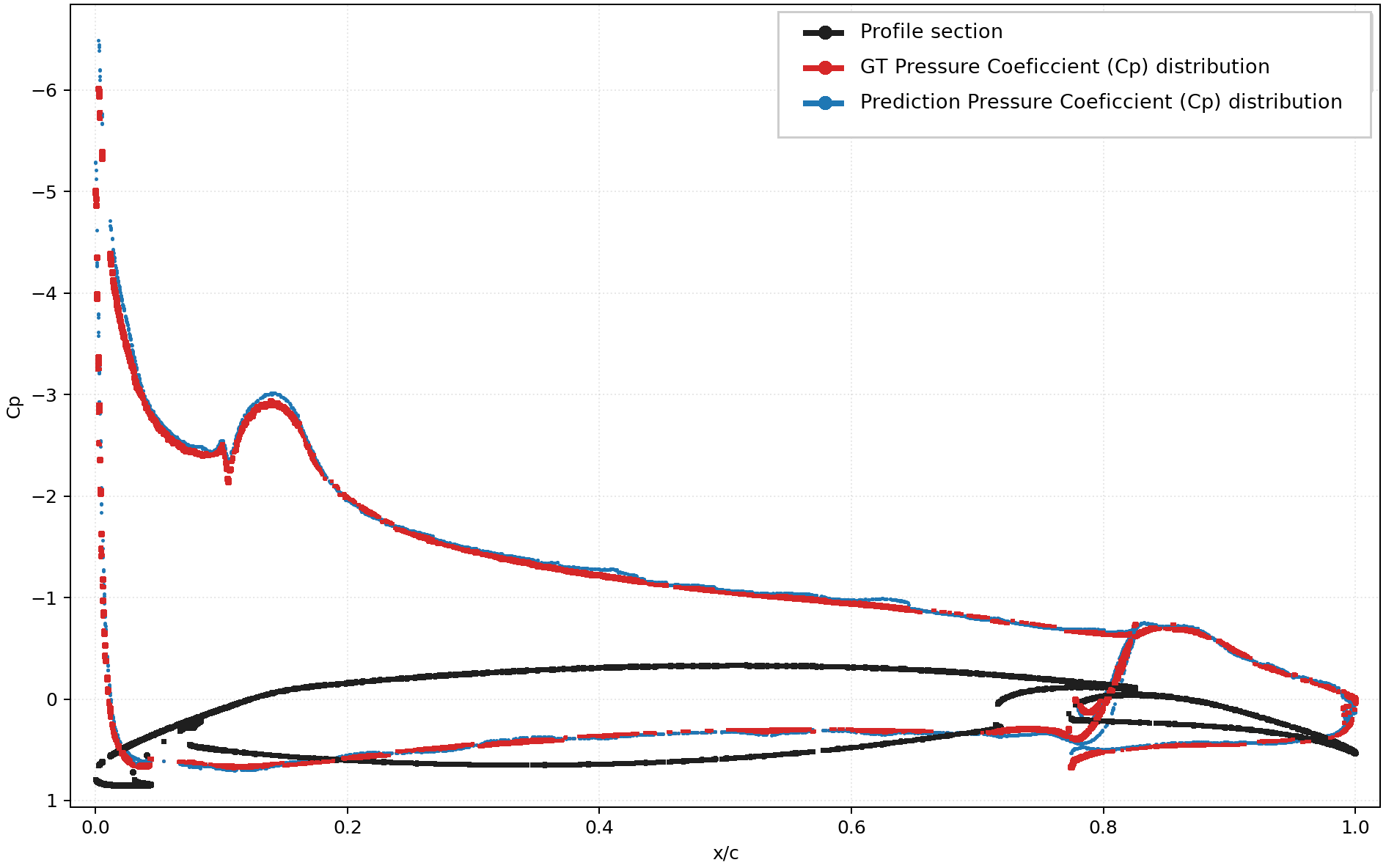}
    \caption{Pressure-coefficient profile extracted from the decoded field on a wing section for case LHC019 at $12^{\circ}$ angle of attack. The comparison shows that AeroJEPA preserves the local pressure distribution needed to recover section-level aerodynamic behavior from the decoded solution.}
    \label{fig:highlift_cp_section_appendix}
\end{figure*}

\begin{figure*}[t]
    \centering
    \includegraphics[width=0.72\textwidth]{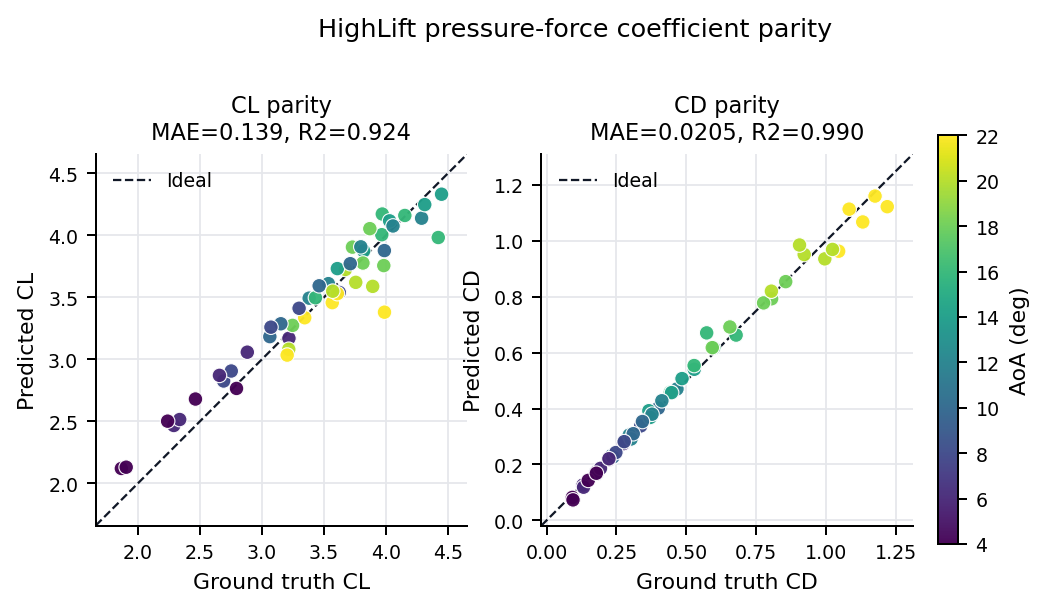}
    \caption{Parity plots for $C_L$ and $C_D$ computed from the decoded predicted fields. Unlike the latent probing experiment, these coefficients are obtained by estimating aerodynamic forces directly from the reconstructed surface solution, showing that AeroJEPA yields accurate integrated force predictions from the decoded field itself.}
    \label{fig:highlift_force_parity_appendix}
\end{figure*}

\paragraph{Decoded-field consistency.}
We also assess whether the decoded fields yield accurate aerodynamic quantities when forces are estimated directly from the reconstructed surface solution, rather than inferred through a linear probe in latent space. Figure~\ref{fig:highlift_cp_section_appendix} shows the pressure-coefficient distribution on a wing section for case LHC019 at $12^{\circ}$ angle of attack, comparing the coefficient extracted from the decoded field against the reference CFD profile. Figure~\ref{fig:highlift_force_parity_appendix} further reports parity plots for $C_L$ and $C_D$ computed from the decoded predicted fields. These results are complementary to the latent probing analysis in the main paper: they show that AeroJEPA not only stores aerodynamically meaningful information in latent space, but also recovers integrated force quantities from the actual decoded flow field with good agreement.

\begin{figure*}[t]
    \centering
    \includegraphics[width=0.82\textwidth]{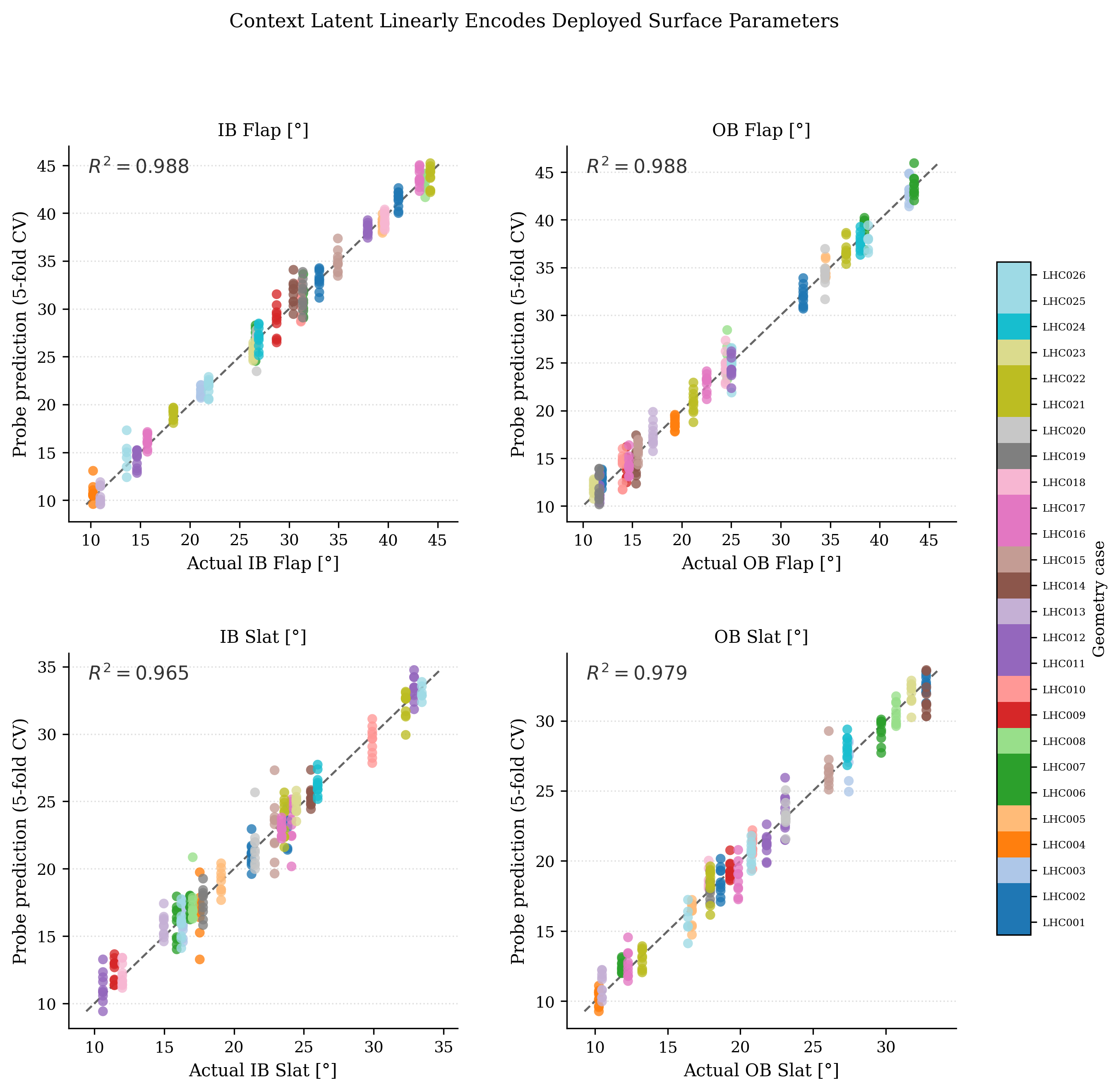}
    \caption{Recovery of HiLift design variables from the context latents using ridge regression. Although the model is trained only from input mesh points and target flow fields, the context representation remains strongly aligned with the underlying geometric parameters.}
    \label{fig:highlift_design_probe_appendix}
\end{figure*}

\paragraph{Linear recovery and concept arithmetic.}
Figure~\ref{fig:highlift_design_probe_appendix} reports the ridge-regression recovery of design variables from the context latents. These quantities are never used during training, so this experiment directly tests whether the context encoder captures geometry information in a form that is linearly accessible. The four dominant control-surface parameters are all recovered with high fidelity ($R^2=0.965$-$0.988$), which supports the interpretation of the context manifold in the main text. Figure~\ref{fig:highlift-arithmetic} complements this result through latent arithmetic. We define concept vectors from the rows of the ridge-regression weight matrix, traverse the context latent space along those directions, and decode the resulting states back into standardized physical design parameters. The targeted variable dominates the response along its corresponding trajectory, while the remaining parameters stay comparatively stable. The residual couplings mainly appear between inboard and outboard flaps and between inboard and outboard slats, which is consistent with the strong geometric dependencies of the high-lift configuration.

{

\subsection{Selective decoding on HiLiftAeroML}
\label{app:selective_decoding}

We evaluate whether a trained AeroJEPA can predict the aerodynamic field on a selected subset of surface coordinates without decoding the remainder of the aircraft. No model is retrained or fine-tuned for this experiment. The evaluation contains 50 held-out cases from five aircraft geometries.

Because the source files do not contain semantic component identifiers, we define a single fixed and visually verified outboard main-wing region using the surface flag and normalized coordinate bounds $-0.55\leq y\leq-0.05$, $-0.55\leq x\leq-0.28$, and $0.44\leq z\leq0.68$. This region excludes the wing root, engine, fuselage, and tail. For every case, we sample $65{,}536$ query points uniformly without replacement from this region. All models are evaluated on exactly the same coordinates and reference CFD values.

Figure~\ref{fig:hilift_selective_decoding} visualizes the absolute pressure error on a representative held-out case. AeroJEPA encodes $131{,}072$ uniformly sampled geometry points from the aircraft surface, as in training, but sends only the selected wing coordinates to its decoder. Predictions are evaluated using query batches of $8{,}192$ points and then repeated with batches of $4{,}096$ points. Transolver and GeoTransolver receive the same $65{,}536$ wing coordinates as two randomly partitioned sets of $32{,}768$ points. A second independent partition is used to measure whether the prediction at a coordinate changes with the other coordinates processed alongside it.

\begin{table*}[t]
\centering
\caption{Selective decoding on the outboard-wing region of 50 held-out HiLiftAeroML cases. Pressure RMSE is reported in the dataset pressure units. Repartition NRMSE measures the change in the four normalized output channels when the same coordinates are grouped into different query sets. Bracketed values are 95\% geometry-cluster bootstrap intervals.}
\label{tab:hilift_selective_decoding}
\resizebox{\textwidth}{!}{%
\begin{tabular}{@{}lcccccc@{}}
\toprule
Model &
Pressure RMSE $\downarrow$ &
Pressure $R^2$ $\uparrow$ &
$u$ $R^2$ $\uparrow$ &
$v$ $R^2$ $\uparrow$ &
$w$ $R^2$ $\uparrow$ &
Repartition NRMSE $\downarrow$ \\
\midrule
AeroJEPA
& \textbf{1.468} [1.080, 1.761]
& \textbf{0.953}
& \textbf{0.932}
& \textbf{0.933}
& \textbf{0.962}
& \textbf{0.0000} \\
Transolver
& 8.462 [7.999, 8.925]
& $-0.161$
& $-0.130$
& $-0.188$
& $-0.064$
& 0.0225 \\
GeoTransolver
& 8.537 [8.017, 9.037]
& $-0.089$
& $-0.094$
& $-0.189$
& $-0.113$
& 0.0134 \\
\bottomrule
\end{tabular}%
}
\end{table*}

Table~\ref{tab:hilift_selective_decoding} summarizes the quantitative selective-decoding results. AeroJEPA obtains a pressure RMSE of $1.468$ and pressure $R^2$ of $0.953$, with similarly high $R^2$ values across the three velocity components. Its predictions are identical under the two query batch sizes, producing a repartition NRMSE of zero. Transolver and GeoTransolver obtain pressure RMSEs of $8.462$ and $8.537$, respectively, and their predictions change when the same coordinates are repartitioned. The experiment therefore demonstrates that AeroJEPA can reuse a geometry encoding to decode a selected spatial region and that its output at a coordinate is independent of the other decoder queries evaluated in the same batch.

\begin{figure*}[t]
\centering
\includegraphics[width=\textwidth]{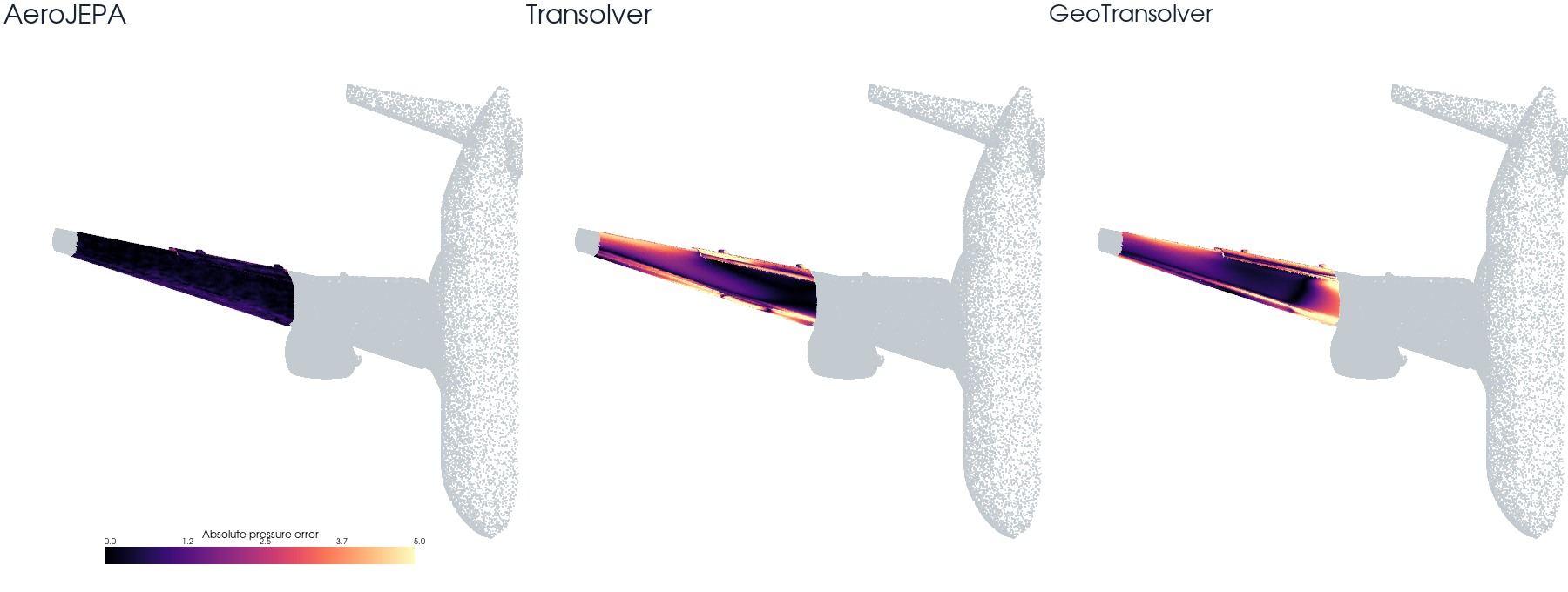}
\caption{Absolute pressure error on the selected outboard-wing region for a representative held-out HiLiftAeroML case. All methods are evaluated on the same wing coordinates. The uncolored remainder of the aircraft shows the geometry and is not included in the displayed field predictions. AeroJEPA maintains substantially lower error over the selected region than the evaluated direct-regression models.}
\label{fig:hilift_selective_decoding}
\end{figure*}

}

\subsection{Additional SuperWing surrogate results}
\label{app:superwing_additional}

\begin{figure*}[t]
    \centering
    \includegraphics[width=0.8\textwidth]{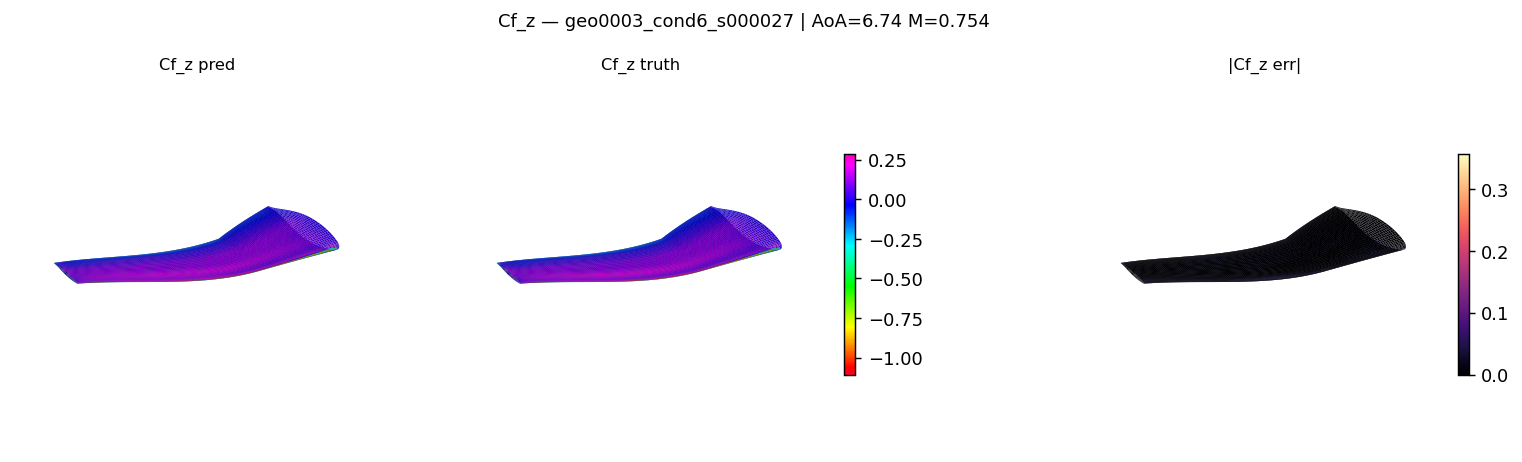}\hfill
    \includegraphics[width=0.8\textwidth]{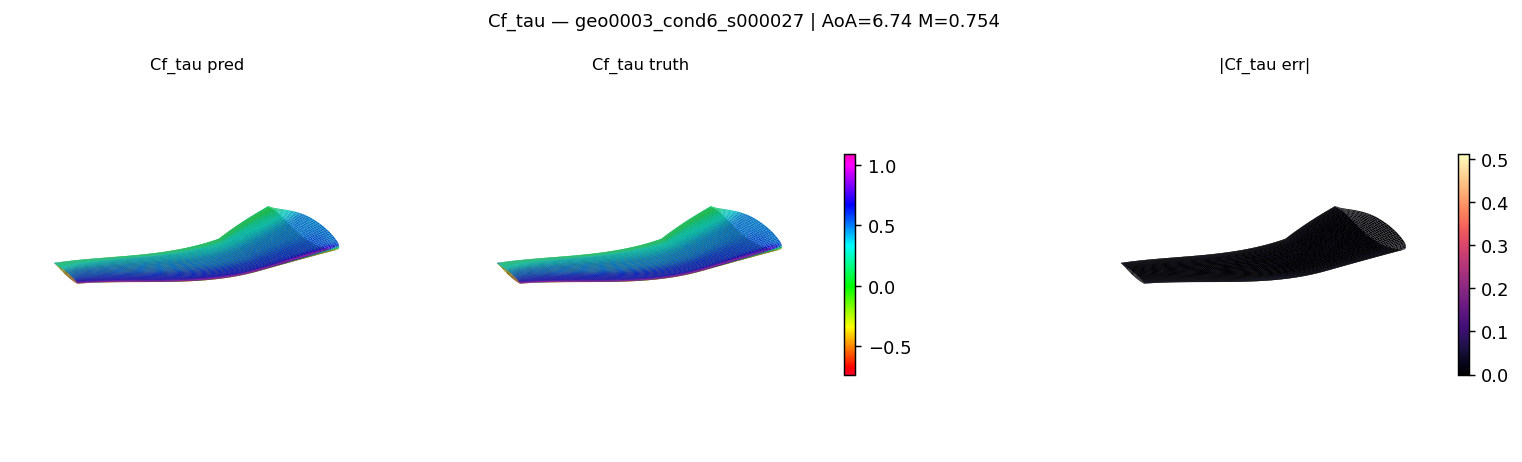}\hfill
    \includegraphics[width=0.8\textwidth]{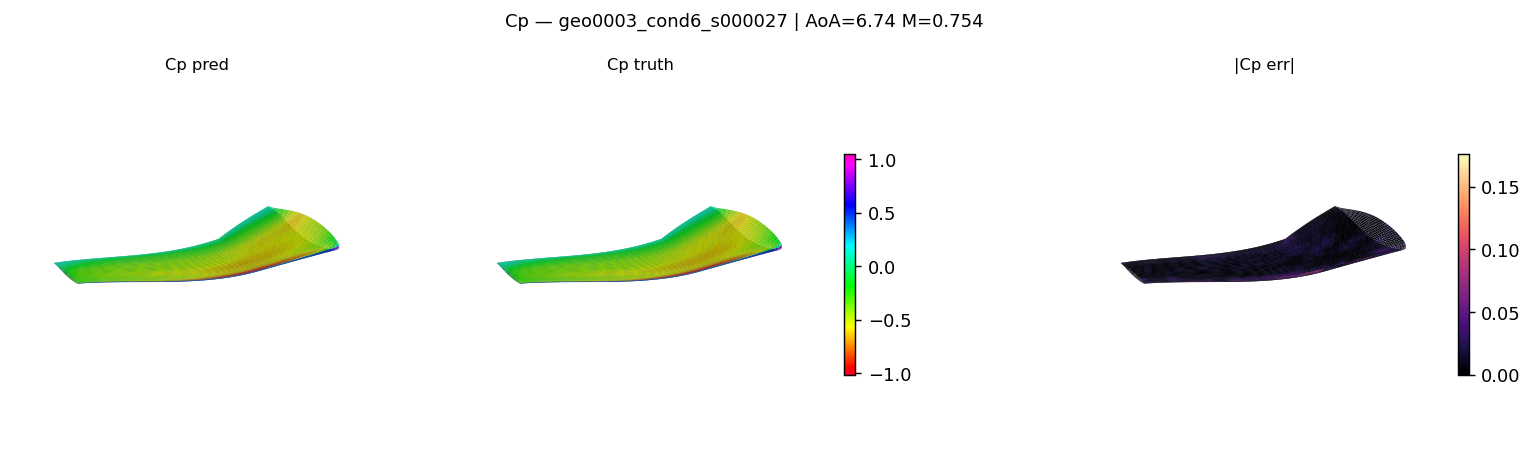}
    \caption{Representative decoded-field comparisons on SuperWing for $C_{f,z}$, $C_{f,\tau}$, and $C_p$. The predicted surface fields remain consistent with the reference solution across the three quantities most relevant to transonic wing analysis.}
    \label{fig:superwing_fields_appendix}
\end{figure*}

\paragraph{Decoded fields.}
Figure~\ref{fig:superwing_fields_appendix} shows representative decoded-field comparisons for $C_{f,z}$, $C_{f,\tau}$, and $C_p$ on a SuperWing test case. The predicted surface fields remain consistent with the reference solution across the three quantities most relevant to transonic wing analysis.

\begin{figure*}[t]
    \centering
    \includegraphics[width=0.68\textwidth]{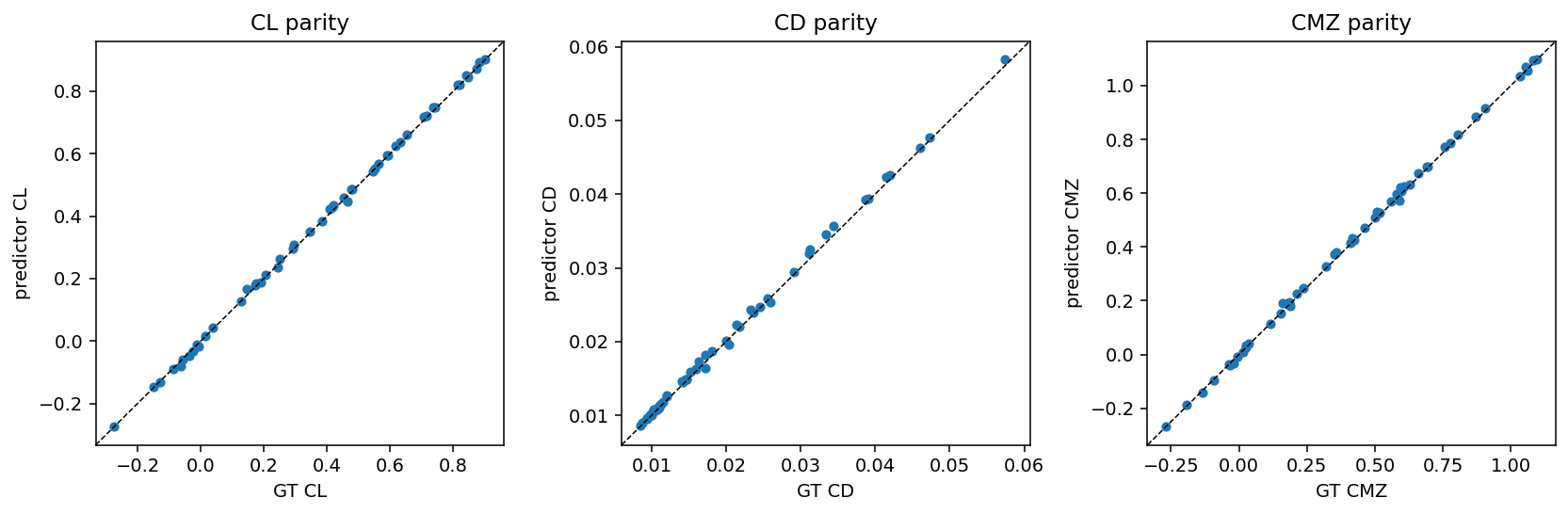}
    \caption{Parity plots for aerodynamic forces estimated from the decoded SuperWing fields. The comparison is performed on $C_L$ and $C_D$ recovered from the reconstructed surface solution, showing that the decoded fields preserve the integrated aerodynamic response with good fidelity.}
    \label{fig:superwing_force_parity_appendix}
\end{figure*}

\paragraph{Integrated aerodynamic response from decoded fields.}
Figure~\ref{fig:superwing_force_parity_appendix} reports parity plots for aerodynamic forces estimated directly from the decoded SuperWing fields. This comparison is performed on $C_L$ and $C_D$ recovered from the reconstructed surface solution itself. The decoded fields therefore preserve not only local surface quantities, but also the integrated aerodynamic response needed for downstream evaluation.

{
\subsection{Training-objective and representation ablations}
\label{app:loss_ablations}

We first study how the representation and reconstruction behavior changes with the training objective. Table~\ref{tab:superwing_loss_ablation} compares three SIGReg weights with a contrastive alternative on SuperWing. Representation quality is evaluated using linear probes trained on the training geometries and evaluated on held-out geometries. Design $R^2$ measures the mean recovery of eight selected design parameters from the context tokens, Aero $R^2$ measures the mean recovery of $C_L$ and $C_D$, and $\Delta$Aero $R^2$ measures recovery of within-geometry changes in these coefficients. The context diagnostics use mean-pooled geometry tokens from held-out cases. Context cosine gap measures the separation between same-geometry, different-condition pairs and condition-matched, different-geometry pairs. Context within variation is the fraction of standardized context-feature variation occurring within rather than between geometries. Effective rank summarizes the spectrum of the centered context covariance.

\begin{table*}[t]
\centering
\caption{Training-objective ablation on SuperWing. Probe metrics are evaluated on held-out geometries. Field MSE is reported in training-normalized units on 12 held-out cases.}
\label{tab:superwing_loss_ablation}
\resizebox{\textwidth}{!}{%
\begin{tabular}{@{}lccccccc@{}}
\toprule
Objective &
Design $R^2$ $\uparrow$ &
Aero $R^2$ $\uparrow$ &
$\Delta$Aero $R^2$ $\uparrow$ &
Context cosine gap $\uparrow$ &
Context within (\%) $\downarrow$ &
Context eff. rank &
Field MSE $\downarrow$ \\
\midrule
SIGReg $\lambda_s=10^{-3}$
& \textbf{0.886}
& \textbf{0.993}
& \textbf{0.987}
& \textbf{0.983}
& 0.94
& 4.78
& \textbf{0.0085} \\
SIGReg $\lambda_s=10^{-2}$
& 0.859
& 0.978
& 0.964
& \textbf{0.983}
& 0.92
& 4.72
& 0.0125 \\
SIGReg $\lambda_s=10^{-1}$
& 0.763
& 0.915
& 0.839
& 0.980
& \textbf{0.91}
& 5.11
& 0.0158 \\
Contrastive
& 0.460
& 0.783
& 0.704
& 0.738
& 17.74
& 3.76
& 0.5072 \\
\bottomrule
\end{tabular}%
}
\end{table*}

The $\lambda_s=10^{-3}$ configuration provides the strongest overall combination of design recovery, aerodynamic recovery, and reconstruction accuracy. Increasing the regularization weight progressively reduces the probe and reconstruction results, while the contrastive alternative substantially degrades both field prediction and the organization of the geometry representation. We therefore use $\lambda_s=10^{-3}$ for the primary AeroJEPA models.

We next compare the complete AeroJEPA objective with reconstruction-only training on HiLiftAeroML in Table~\ref{tab:hilift_reconstruction_ablation}. Both models use the same architecture and are evaluated after 30 epochs, so this experiment isolates the effect of latent matching at an equal training budget. These matched early checkpoints are distinct from the fully trained model used for the primary HiLiftAeroML results.

\begin{table*}[t]
\centering
\caption{Matched 30-epoch comparison between the complete AeroJEPA objective and reconstruction-only training on HiLiftAeroML. Design and aerodynamic quantities are recovered using linear probes.}
\label{tab:hilift_reconstruction_ablation}
\resizebox{\textwidth}{!}{%
\begin{tabular}{@{}lcccccccccc@{}}
\toprule
Objective &
IB flap $R^2$ $\uparrow$ &
OB flap $R^2$ $\uparrow$ &
IB slat $R^2$ $\uparrow$ &
OB slat $R^2$ $\uparrow$ &
$\Delta C_L$ $R^2$ $\uparrow$ &
$\Delta C_D$ $R^2$ $\uparrow$ &
Context gap $\uparrow$ &
Context R@1 $\uparrow$ &
Predicted R@1 $\uparrow$ &
Eff. rank \\
\midrule
Full AeroJEPA
& \textbf{0.948}
& \textbf{0.962}
& \textbf{0.839}
& 0.596
& \textbf{0.822}
& \textbf{0.661}
& \textbf{1.341}
& \textbf{100\%}
& \textbf{92\%}
& \textbf{4.68} \\
Reconstruction only
& 0.886
& 0.873
& 0.621
& \textbf{0.739}
& 0.736
& 0.621
& 0.800
& 88\%
& 4\%
& 1.76 \\
\bottomrule
\end{tabular}%
}
\end{table*}

Reconstruction-only training retains substantial linearly accessible information, confirming that the architecture itself provides a strong inductive bias. The complete objective nevertheless improves three of the four control-surface probes, both aerodynamic-change probes, and the organization of both latent spaces. The largest difference appears in predicted-token retrieval, which increases from $4\%$ to $92\%$, while context retrieval increases from $88\%$ to $100\%$. These results show that latent matching provides a substantially clearer organization of the predicted aerodynamic representation while preserving reconstruction capability.
}

{

\subsection{Latent-space aerodynamic optimization}
\label{app:latent_optimization}

We evaluate whether the learned geometry representation can serve as a differentiable search space for preliminary aerodynamic design exploration. On SuperWing, we maximize aerodynamic efficiency $C_L/C_D$ at a fixed cruise condition $(\alpha_{\mathrm{cruise}},M_{\mathrm{cruise}})$ by treating the $128$-dimensional mean-pooled context representation $z_{\mathrm{ctx}}$ as the optimization variable. The search uses neither CAD modification nor field decoding inside the optimization loop.

\paragraph{Differentiable objective.}
The frozen AeroJEPA predictor maps the context representation and operating condition to a predicted aerodynamic representation,
\begin{equation}
z_{\mathrm{pred}}
=
\Phi_{\mathrm{pred}}
\left(
z_{\mathrm{ctx}},
\alpha_{\mathrm{cruise}},
M_{\mathrm{cruise}}
\right).
\end{equation}
Linear ridge probes, defined in Appendix~\ref{app:linear_probes}, then recover $C_L$ and $C_D$ from $z_{\mathrm{pred}}$. Their composition produces an end-to-end differentiable map from geometry representation to aerodynamic coefficients. We optimize
\begin{equation}
\min_{z_{\mathrm{ctx}}\in\mathbb{R}^{128}}
J(z_{\mathrm{ctx}})
=
-\,\frac{
C_L(z_{\mathrm{ctx}})
}{
C_D(z_{\mathrm{ctx}})
},
\label{eq:latent_optimization_objective}
\end{equation}
with gradients obtained by backpropagation through the frozen predictor and probes.

\paragraph{Search constraints.}
Unconstrained optimization could move the representation outside the region supported by the training data. We therefore solve Equation~\ref{eq:latent_optimization_objective} using SLSQP subject to four families of constraints derived from training-set statistics. First, a Mahalanobis trust region restricts the search around the training distribution,
\begin{equation}
\left(
z_{\mathrm{ctx}}-\mu_{\mathrm{train}}
\right)^\top
\Sigma_{\mathrm{train}}^{-1}
\left(
z_{\mathrm{ctx}}-\mu_{\mathrm{train}}
\right)
\leq \tau.
\end{equation}
Second, we constrain the nine SuperWing design parameters that are reliably recovered from the context representation, selected using five-fold cross-validated $R^2\geq0.85$,
\begin{equation}
x_k^{\min}
\leq
W_k^\top\widetilde{z}_{\mathrm{ctx}}+b_k
\leq
x_k^{\max}.
\end{equation}
Third, we impose a drag floor $C_D\geq0.9\min_{\mathrm{train}}C_D$ and lift and drag ceilings equal to $1.05$ times their respective training-set maxima. Finally, the predicted $L/D$ is bounded by the maximum observed in the dataset. Together, these constraints restrict the search to a calibrated region of the geometry and aerodynamic representations.

We perform eight random restarts. Initial context representations are drawn from the training distribution and projected inside the Mahalanobis trust region. Jacobians for the affine design constraints and Mahalanobis constraint are supplied analytically, while the aerodynamic gradients are obtained using automatic differentiation. Once SLSQP converges to $z_{\mathrm{ctx}}^\star$, the corresponding design vector is recovered using the linear design probe,
\begin{equation}
x^\star
=
W_{\mathrm{design}}
\widetilde{z}_{\mathrm{ctx}}^\star
+
b_{\mathrm{design}}.
\end{equation}
We then retrieve the dataset geometry whose standardized design vector is closest to $x^\star$ in Euclidean distance.

\paragraph{Optimization behavior.}
All eight restarts converge to the same neighborhood while satisfying the imposed constraints. Figure~\ref{fig:latent_trajectory} shows the optimization trajectories in a PCA projection of the training context representations. The SLSQP iterations move toward larger probe-predicted $C_L/C_D$ values while remaining inside the trust region. The final latent solution lies close to its retrieved dataset geometry, indicating that the optimization terminates near a region represented by observed wings.

\begin{figure}[t]
\centering
\includegraphics[width=0.86\linewidth]{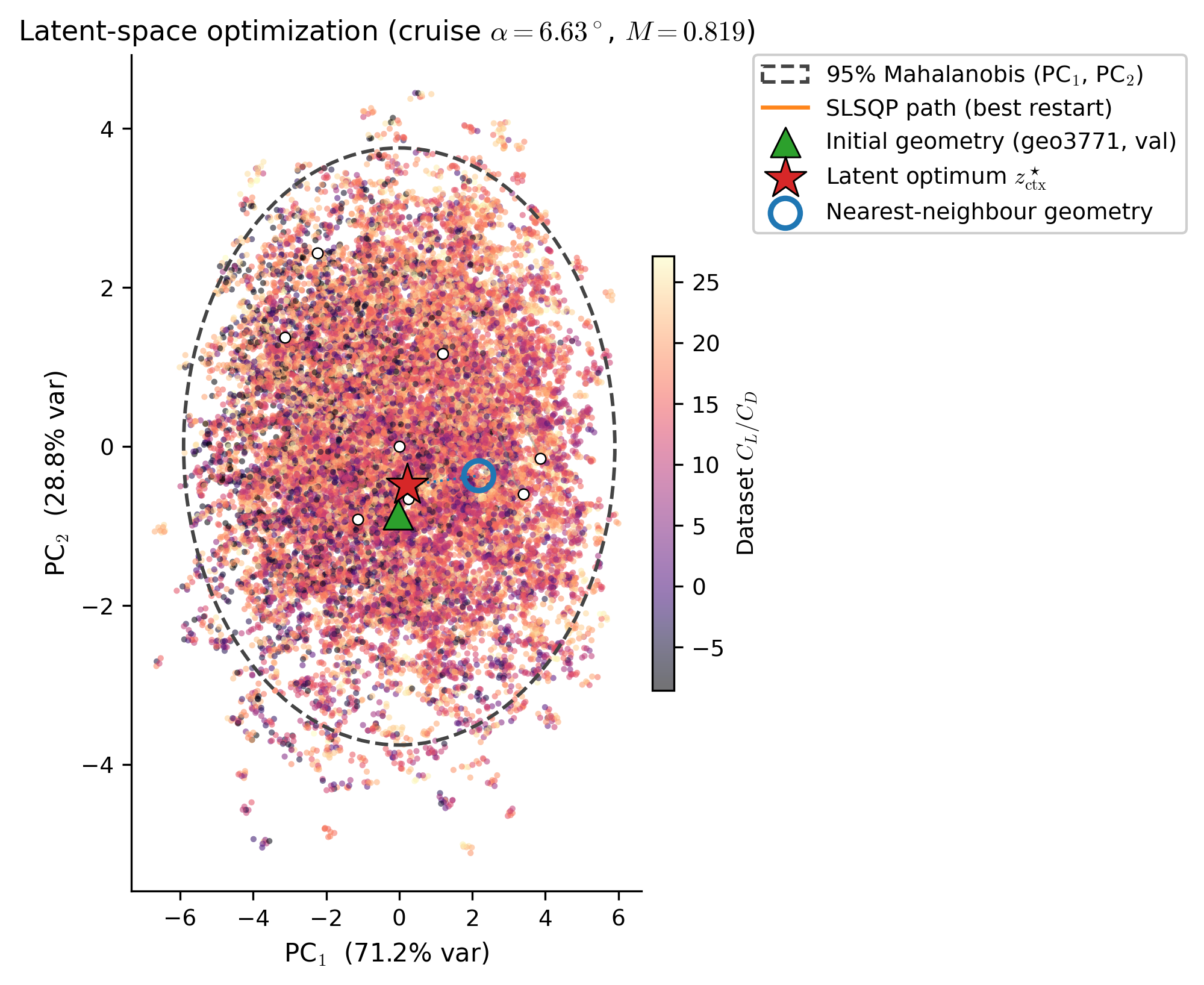}
\caption{Latent-space optimization trajectories. The training context representations are projected onto two principal components and colored by dataset $C_L/C_D$. Gray curves show the eight SLSQP restarts, and the orange curve highlights the restart producing the selected optimum. White circles indicate initial points, the red star indicates $z_{\mathrm{ctx}}^\star$, and the blue circle indicates the dataset geometry retrieved as its nearest neighbor in design space. The dashed contour shows the projected Mahalanobis trust region.}
\label{fig:latent_trajectory}
\end{figure}

Figure~\ref{fig:superwing_optimization} relates the optimized representation to the aerodynamic envelope and design parameters. The predicted optimum moves from its initialization toward the high-efficiency region of the observed $(C_D,C_L)$ distribution while satisfying the imposed lift, drag, and $L/D$ constraints. The recovered design combines a large reference area and aspect ratio, high sweep, aggressive taper, and root-biased washout. Its nearest observed neighbor follows a similar profile over the nine reliably recovered design parameters, connecting the optimized representation to a wing configuration present in the dataset.

\begin{figure*}[t]
\centering
\includegraphics[width=0.59\textwidth]{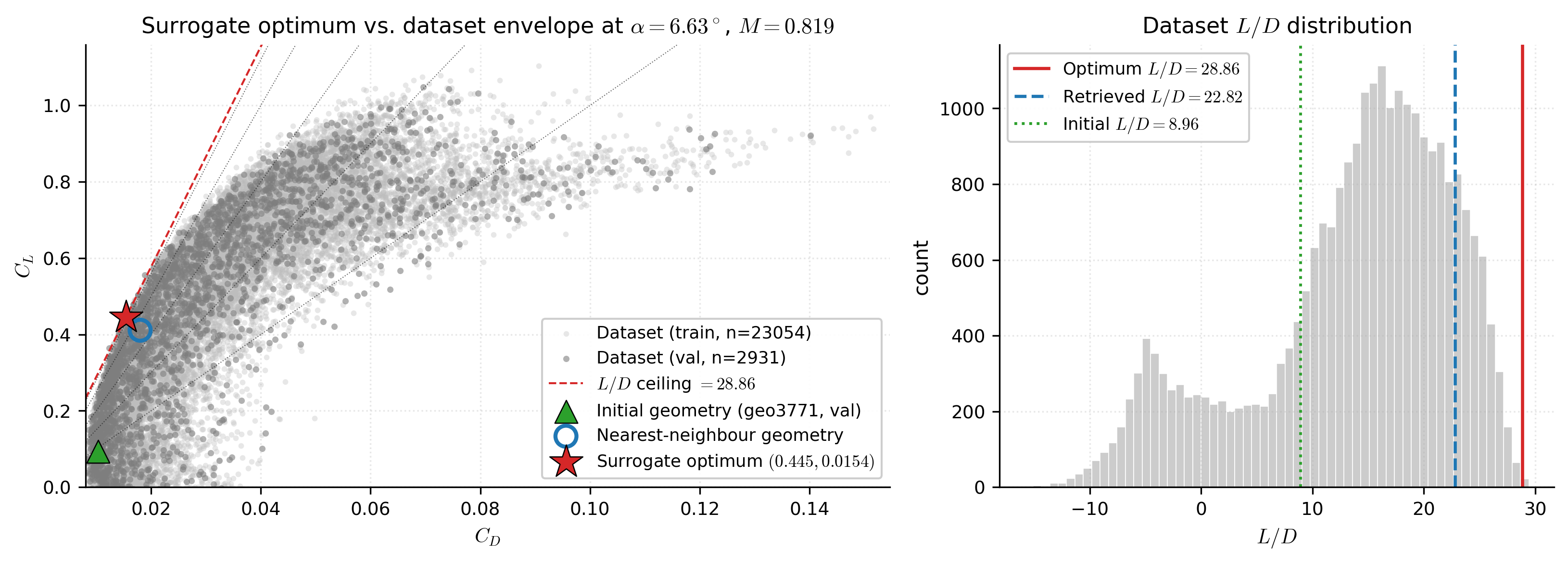}\hfill
\includegraphics[width=0.39\textwidth]{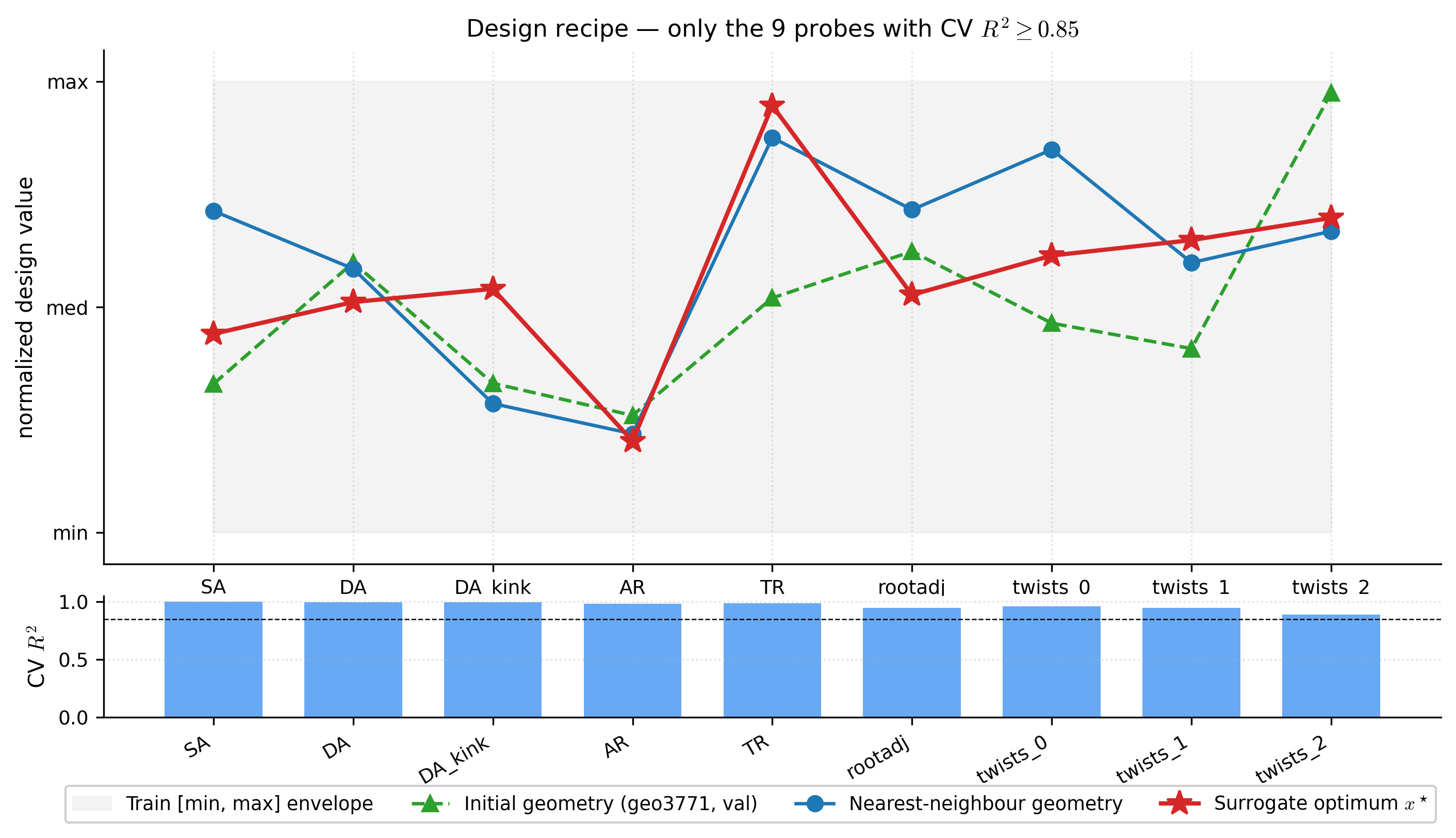}
\caption{Constrained latent-space optimization on SuperWing. \emph{Left:} the probe-predicted optimum approaches the high-efficiency region of the observed $(C_D,C_L)$ envelope and lies near the right tail of the dataset $L/D$ distribution. \emph{Right:} parallel-coordinates comparison of the initial design, the recovered latent optimum, and its nearest observed wing over the nine reliably decoded design parameters. The nearest observed wing follows a similar parameter profile to the optimized latent point.}
\label{fig:superwing_optimization}
\end{figure*}

The experiment shows that the context representation provides a smooth and differentiable search space, that its linear design and aerodynamic readouts can support constrained optimization, and that the resulting point can be translated into an interpretable design parameterization. The optimization requires only the frozen predictor and lightweight linear probes, avoiding repeated field decoding during the search. Because the synthetic optimum has not been evaluated through a new CFD simulation, this experiment establishes the feasibility of representation-based design exploration rather than verified aerodynamic improvement.

}

\subsection{Concept-vector latent arithmetic on HiLiftAeroML}
\label{app:highlift-arithmetic}

A semantic latent space should not only support gradient-based optimization
(App.~\ref{app:latent_optimization}) but also expose interpretable
\emph{directions} along which design intent can be expressed by simple
vector arithmetic. We test this on HiLiftAeroML by asking whether
the AeroJEPA's $64$-dimensional (mean) context latent $z_\mathrm{ctx}$ encodes
high-lift control surfaces -- inboard / outboard flap and slat deflections --
along separable linear axes, despite never having been trained on those
labels.

\paragraph{Concept directions from linear probes.}
We fit a single ridge probe per design parameter
$x_k$ on the train split,
$x_k = w_k^\top \tilde z_\mathrm{ctx} + b_k$ with
$\tilde z_\mathrm{ctx} = (z_\mathrm{ctx} - \mu_\mathrm{ctx})/\sigma_\mathrm{ctx}$,
where the design parameters are the four main HiLiftAeroML deflections
(IB Flap, OB Flap, IB Slat, OB Slat) plus the two flap-gap multipliers.
The unit-norm probe direction $v_k = w_k / \lVert w_k \rVert$ is interpreted
as the latent ``concept vector'' for parameter $x_k$. To probe the
disentanglement of the representation, we walk the train-mean latent along
one direction at a time,
\begin{equation}
z_\mathrm{ctx}(\gamma) \;=\; \mu_\mathrm{ctx} + \gamma\, v_k,
\end{equation}
and at each step $\gamma$ predict \emph{all} design parameters from the
shifted latent through the same probe matrix. The slope of the predicted
parameter $x_j$ along the walk on direction $v_k$ has a closed form,
\begin{equation}
\frac{\mathrm{d}\tilde x_j}{\mathrm{d}\gamma}
= \frac{\bigl(W_\mathrm{ctx}\,(v_k / \sigma_\mathrm{ctx})\bigr)_j}{\sigma_{x_j}},
\end{equation}
expressed in train-set standard deviations of $x_j$ per unit step in latent
space; we report this number as the panel-level
``$\sigma/\gamma$'' sensitivity.

\begin{figure}[t]
\centering
\includegraphics[width=\linewidth]{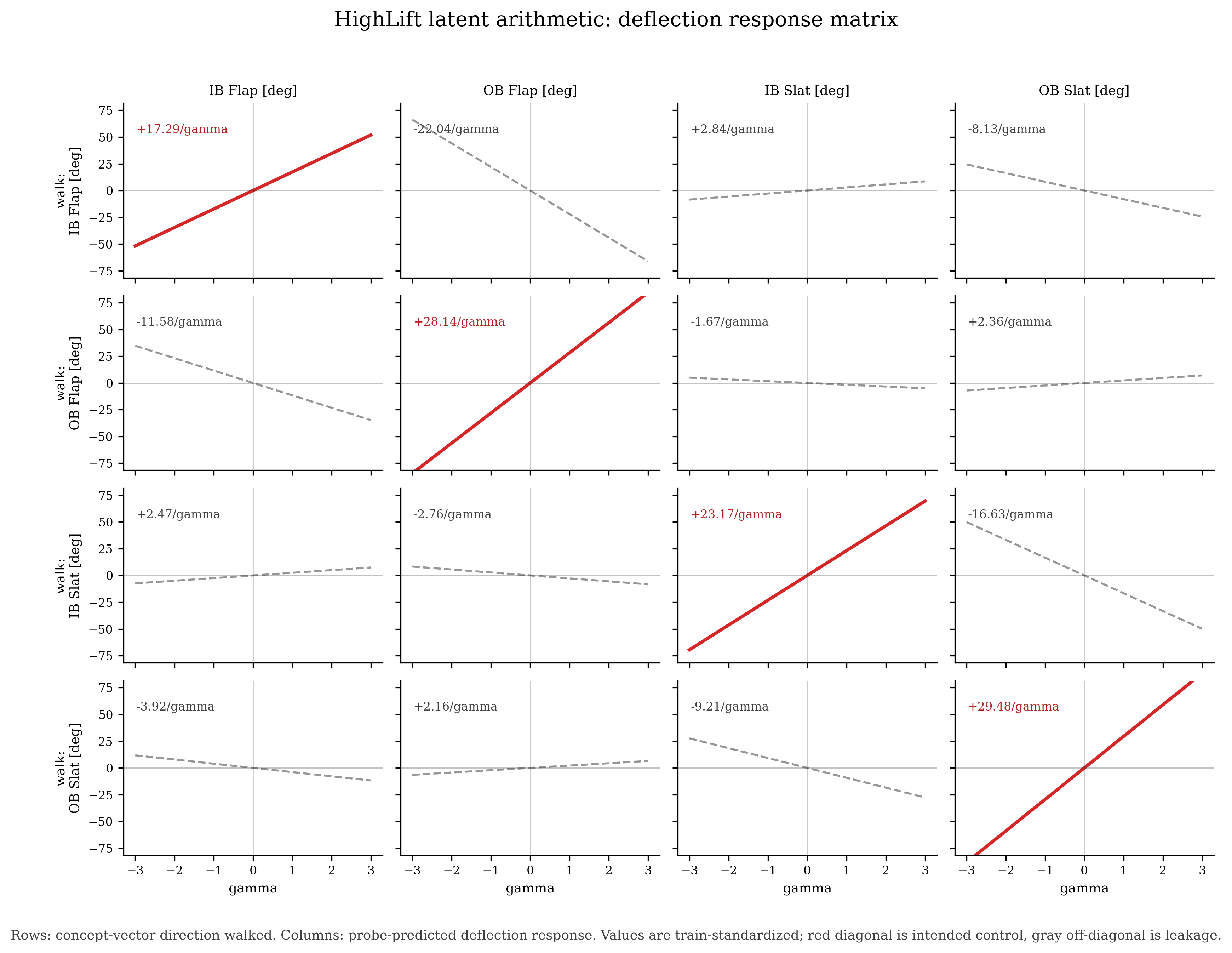}
\caption{%
\textbf{Concept-vector arithmetic in the HiLiftAeroML context latent.}
$4 \times 4$ response matrix obtained by walking the train-mean latent
$\mu_\mathrm{ctx}$ along the unit-norm linear-probe direction of one
design parameter at a time. \emph{Rows:} latent direction walked
(IB Flap, OB Flap, IB Slat, OB Slat). \emph{Columns:} design parameter
read out by the corresponding linear probe. Each panel reports the
sensitivity in train standard deviations per unit walk
($\sigma / \gamma$). Diagonal panels (red) show the intended concept
response and reach near-unit slope. Off-diagonal panels (grey, dashed)
quantify cross-talk between concept directions: the flap--slat
off-diagonal block is essentially flat, indicating limited cross-talk between the flap and slat directions, while inboard / outboard
pairs of the same control surface partially co-activate -- the same
correlation engineers impose by deflecting IB and OB segments in tandem.
The disentanglement is obtained \emph{without any design-parameter
supervision} during AeroJEPA training; the linear probes are fit only on
the train split.
}
\label{fig:highlift-arithmetic}
\end{figure}

\paragraph{The disentanglement matrix.}
Figure~\ref{fig:highlift-arithmetic} shows the resulting $4\times 4$
matrix for the four main deflections. Rows index the latent direction
walked; columns index the design parameter read out. The diagonal panels
(highlighted in red) measure the \emph{intended} response of each concept
direction: walking the IB-Flap concept vector should change predicted
IB-Flap deflection. The off-diagonal panels measure \emph{cross-talk}: how
much an unrelated parameter moves as a side-effect.

The structure in Figure~\ref{fig:highlift-arithmetic} shows that each main deflection has a strongly aligned latent direction. The diagonal responses are substantially larger than most off-diagonal responses, while flap-to-slat and slat-to-flap interactions remain small. The main residual couplings occur between the inboard and outboard elements of the same control surface. Thus, the representation separates flap and slat variations while retaining the correlations between related inboard and outboard elements observed in the dataset.

The probes used to define every concept direction are linear and trained
\emph{only} on the train split, with no design-parameter supervision ever
seen by the AeroJEPA itself. Yet in the resulting representation, design
intent maps onto linear directions, and the residual entanglement is not
arbitrary noise but the physically meaningful coupling that engineers
already build into multi-element high-lift systems. Three points follow:
\begin{enumerate}
\item The geometry-only context encoder \emph{discovers} the
underlying parametric basis of the design space without being told what it
is. This is the practical signal that the latent space is genuinely
semantic rather than merely high-capacity.
\item Concept-vector arithmetic provides a cheap interface for exploring
configurations: a designer can push along the OB-Slat direction and read
the predicted flap and slat values, getting an immediate, calibrated
preview of what changes co-occur in real wings.
\item The \emph{block-diagonal disentanglement structure} between flap and
slat axes is a constructive ingredient for the optimization in
App.~\ref{app:latent_optimization}: when a constrained search needs to
move design parameters independently, the latent space provides
near-orthogonal directions to do so along the parameters that are
geometrically separable, while preserving the natural coupling between IB
and OB segments where the dataset itself shows it.
\end{enumerate}
Together with the gradient-based optimization of
App.~\ref{app:latent_optimization}, this experiment characterizes the AeroJEPA
latent both as a \emph{search space} (smooth, gradient-friendly, linearly
decodable) and as an \emph{interpretation space} (separable concept
directions reflecting physical actuation modes).

\subsection{Linear-probe analysis of the AeroJEPA latent space}
\label{app:linear_probes}

The aerodynamic optimization in App.~\ref{app:latent_optimization} and the
concept-vector arithmetic in App.~\ref{app:highlift-arithmetic} both rely
on a single, simple primitive: a \emph{linear} readout from the AeroJEPA representations. This section defines that readout, lists the
three latent quantities it is applied to, and reports the
goodness-of-fit numbers that justify treating those readouts as semantic.

\paragraph{Probe definition.}
For a latent vector $z \in \mathbb{R}^{D}$ and a scalar target $y$, we
define a \emph{linear probe} as the composition of train-set
standardisation and ridge regression with cross-validated regularisation,
\begin{equation}
\hat y(z) \;=\; w^\top \tilde z + b,
\qquad
\tilde z \;=\; \frac{z - \mu_\mathrm{train}}{\sigma_\mathrm{train}},
\label{eq:probe}
\end{equation}
where $(w, b)$ are obtained by minimising
$\lVert \tilde Z w + b\,\mathbf{1} - y \rVert_2^{2}
 + \lambda\,\lVert w \rVert_2^{2}$
on the train split, with $\lambda$ selected by an inner cross-validation
over $\lambda \in \{10^{-4}, 10^{-3}, \dots, 10^{4}\}$.

\paragraph{Three latent quantities, three probe families.}
We extract three mean-pooled latent representations per case, with dimension $D=64$ on HiLiftAeroML and $D=128$ on SuperWing:
\begin{itemize}
\item \emph{Context representation} $z_{\mathrm{ctx}}\in\mathbb{R}^{D}$, obtained by mean-pooling the context tokens produced from the geometry, before conditioning on the operating variables.
\item \emph{Predicted representation} $z_{\mathrm{pred}}\in\mathbb{R}^{D}$, obtained by mean-pooling the predictor output after conditioning the geometry representation on the operating variables.
\item \emph{Target representation} $z_{\mathrm{tgt}}\in\mathbb{R}^{D}$, obtained by mean-pooling the target-encoder tokens produced from the reference flow field. It provides the latent prediction target during training and is discarded at inference.
\end{itemize}
The following three probe families evaluate which quantities are linearly accessible from the context and predicted representations:
\begin{enumerate}
\item \textbf{Context $\to$ design parameters.} For each design parameter
$x_k$ (e.g.\ each of the $54$ SuperWing morphological coefficients or the
$8$ HiLiftAeroML control-surface deflections), a separate probe with
$z = z_\mathrm{ctx}$ and $y = x_k$. The full set of $K$ probes assembles
into a single matrix readout
$x \approx W_\mathrm{ctx}\,\tilde z_\mathrm{ctx} + b_\mathrm{ctx}$
because every probe shares the same standardisation
$(\mu_\mathrm{ctx}, \sigma_\mathrm{ctx})$.
\item \textbf{Predicted $\to$ aerodynamic coefficients.} Two probes,
$z = z_\mathrm{pred}$, $y \in \{C_L, C_D\}$.
\item \textbf{Predicted $\to$ AoA (control).} A single probe with
$z = z_\mathrm{pred}$ and $y = \alpha$, used to verify that the predictor
output -- but not the geometry-only context -- carries flow-condition
information.
\end{enumerate}

\paragraph{What the probes report on each dataset.}
On SuperWing, the primary configuration achieves a mean predicted-representation $R^2$ of $0.993$ for $C_L$ and $C_D$, and a mean $R^2$ of $0.987$ for their within-geometry changes, as reported in Appendix~\ref{app:loss_ablations}. Eight selected design parameters are recovered from the predicted representation with a mean $R^2$ of $0.886$. Separately, nine of the 54 SuperWing design parameters satisfy the cross-validated threshold $R^2\geq0.85$ when decoded from the context representation, and these parameters define the design constraints used in Appendix~\ref{app:latent_optimization}. On the fully trained HiLiftAeroML model, the four dominant control-surface deflections are recovered from the context representation with $R^2$ values between $0.965$ and $0.988$, while the predicted representation supports linear recovery of $C_L$ and $C_D$ with $R^2=0.930$ and $0.996$, respectively.

\paragraph{From probes to a downstream tool.}
Equation~\eqref{eq:probe} has two properties that the rest of the paper
exploits. First, it is \emph{closed-form differentiable} in $z$ -- the
gradient $\partial \hat y / \partial z = w / \sigma_\mathrm{train}$ is
just a rescaled probe vector. This is what allows the SLSQP optimization
of App.~\ref{app:latent_optimization} to back-propagate through the
predictor and the probes in a single autograd graph, with the
design-bound constraints expressed as affine inequalities with closed-form
Jacobians. Second, the unit-norm probe direction
$v_k = w_k / \lVert w_k \rVert$ is interpreted as the latent
\emph{concept vector} for parameter $x_k$; walking $z_\mathrm{ctx}$ along
$v_k$ and reading out every other parameter gives the disentanglement
matrix of App.~\ref{app:highlift-arithmetic}.

The probes are the bridge between the learned
representation and the design-space quantities a domain expert cares
about. They turn the AeroJEPA latents into a \emph{usable} object -- one that
can be searched (App.~\ref{app:latent_optimization}) and queried by
analogy (App.~\ref{app:highlift-arithmetic}) - without ever fine-tuning
the underlying network.

\end{document}

\end{document}